\documentclass[12pt,authoryear]{elsarticle}
\usepackage{color}
\usepackage{amsmath}
\usepackage{amssymb}
\usepackage{bm}
\usepackage{url}
\usepackage[colorlinks=true]{hyperref}
\usepackage{multirow}
\usepackage{booktabs}
\usepackage{multirow}
\usepackage[figuresright]{rotating}
\usepackage{oubraces}
\usepackage{mathtools}
\usepackage{subfigure}
\usepackage{rotating}
\usepackage{tabu}
\usepackage{longtable}

\DeclareMathSymbol{\R}{\mathalpha}{AMSb}{"52}

\let\today\relax
\makeatletter
\def\ps@pprintTitle{%
    \let\@oddhead\@empty
    \let\@evenhead\@empty
    \def\@oddfoot{\footnotesize\itshape
         {Published in \textbf{Neural Networks} (editorial preprint)} \hfill\today}%
    \let\@evenfoot\@oddfoot
    }
\makeatother

\definecolor{bostonuniversityred}{rgb}{0.0, 0.0, 0.0}

\journal{Neural Networks}

\begin{document}

\begin{frontmatter}

\title{Missing Data Imputation with Adversarially-trained Graph Convolutional Networks}

\author[sapienza]{Indro Spinelli}

\author[sapienza]{Simone Scardapane\corref{cor1}}
\ead{simone.scardapane@uniroma1.it}
\cortext[cor1]{Corresponding author. Phone: +39 06 44585495, Fax: +39 06 4873300.}

\author[sapienza]{Aurelio Uncini}

\address[sapienza]{Department of Information Engineering, Electronics and Telecommunications (DIET), Sapienza University of Rome, Via Eudossiana 18, 00184 Rome, Italy}

\begin{abstract}
Missing data imputation (MDI) is the task of replacing missing values in a dataset with alternative, predicted ones. Because of the widespread presence of missing data, it is a fundamental problem in many scientific disciplines. Popular methods for MDI use global statistics computed from the entire dataset (e.g., the feature-wise medians), or build predictive models operating independently on every instance. In this paper we propose a more general framework for MDI, leveraging recent work in the field of graph neural networks (GNNs). We formulate the MDI task in terms of a graph denoising autoencoder, where each edge of the graph encodes the similarity between two patterns. A GNN encoder learns to build intermediate representations for each example by interleaving classical projection layers and locally combining information between neighbors, while another decoding GNN learns to reconstruct the full imputed dataset from this intermediate embedding. In order to speed-up training and improve the performance, we use a combination of multiple losses, including an adversarial loss implemented with the Wasserstein metric and a gradient penalty. We also explore a few extensions to the basic architecture involving the use of residual connections between layers, and of global statistics computed from the dataset to improve the accuracy. {\color{bostonuniversityred}On a large experimental evaluation with varying levels of artificial noise, we show that our method is on par or better than several alternative imputation methods. On three datasets with pre-existing missing values, we show that our method is robust to the choice of a downstream classifier, obtaining similar or slightly higher results compared to other choices.}
\end{abstract}


\end{frontmatter}

\section{Introduction}
\label{sec:introduction}

While machine learning and deep learning have achieved tremendous results over the last years \citep{goodfellow2016deep}, with new breaktroughs arising constantly (e.g., in drug discovery \citep{chen2018rise}), the vast majority of supervised learning methods still require datasets with complete information. At the same time, many real-world problems require dealing with incomplete data, such as in the biomedical or insurance sectors \citep{van2018flexible}. For this reason, flexible missing data imputation (MDI) methods are a fundamental component for widespread adoption of machine learning. An MDI algorithm takes a dataset with missing values in some of its input vectors, and replaces these values with some appropriately predicted ones in order to obtain a full dataset.\footnote{In the literature, this problem is also called missing value imputation (MVI) \citep{lin2019missing}. In this paper, we use the terms \textit{data} and \textit{value} interchangeably based on context.} In particular, there is the need for powerful \textit{multivariate} imputation methods able to work in a variety of data generation regimes \citep{yoon2018gain}.

It has been recognized for a while that data imputation can be framed under a predictive framework, and classical machine learning methods (e.g., for regression and classification) might be adapted for this task \citep{bertsimas2017predictive}. However, some care must be taken when adapting them, for two fundamental reasons. Firstly, different inputs in general have different missing components, while most machine learning models assume full input vectors. Secondly, MDI might be a simple preprocessing step for downstream learning tasks, and in this case performance in terms of reconstruction might not be a perfect proxy for classification/regression accuracy later on.

In general, the resulting predictive approaches to MDI can be classified depending on whether they try to build a global model for data imputation, or whether they use similar data points to infer the missing components. Algorithms in the latter class include using simple statistics computed from the entire dataset (e.g., medians), or more advanced k-NN strategies \citep{lakshminarayan1996imputation}. In the former case, instead, we have simple linear models \citep{lakshminarayan1996imputation}, support vector machines \citep{wang2006missing} or, more recently, deep neural architectures \citep{yoon2018gain}. These are surveyed more in-depth in Section \ref{subsec:missing_data_imputation}. 

We argue that a more powerful technique for MDI should exploit both ideas, i.e., use similar data points for each imputation \textit{and} global models built from the overall dataset. 
In fact, recently a large class of neural network techniques have emerged that are able to model and exploit this kind of structured information (in the form of relationships between examples), by working in the domain of graphs \citep{bronstein2017geometric,battaglia2018relational}. These models have been applied successfully to a wide range of problems, among which recommender systems \citep{ying2018graph}, quantum chemistry \citep{gilmer2017neural}, entity extraction from relational data \citep{schlichtkrull2018modeling}, semi-supervised learning \citep{kipf2016semi}, and many others. However, to the best of our knowledge these techniques have never been applied to MDI. To overcome this, in this paper we define an architecture for MDI based on a specific class of graph neural networks, namely, graph convolutional networks (GCN), and empirically evaluate it on a large set of benchmark datasets \citep{kipf2016semi}.

\subsection*{Contributions of the paper}

Our generic framework for MDI is shown later on in Figure \ref{fig:framework}. We frame the overall problem in terms of a GCN autoencoder,\footnote{We use the term autoencoder to refer to any architecture that learns to map an input (or a corrupted version in the case of denoising autoencoders) to itself.} that learns to reconstruct the overall dataset conditioned on some artificial noise added during the training phase (similar to a classical denoising autoencoder \citep{vincent2008dae}). To build a graph of similarities between points we leverage prior literature on manifold regularization \citep{belkin2006manifold}, and we describe a simple technique that was found to work well in most situations.

After describing the basic architecture, we also detail three extensions to it that are able to improve either the accuracy or the speed of convergence:
\begin{itemize}
    \item Firstly, we train the autoencoder with a mixture of standard loss functions and an adversarial loss, which was shown to provide significant improvements for denoising autoencoders in the non-graph case \citep{yoon2018gain}.
    \item Secondly, we motivate another extension with the inclusion of residual connections from the input to the output layer, similar to residual networks \citep{he2016deep}.
    \item Finally, we also describe how to include global information on the dataset (e.g., means and medians for all feature columns) using a generic context vector in input to the GCN layers.
\end{itemize}

{\color{bostonuniversityred}We test our overall architecture on a large benchmark of datasets with varying levels of artificially-added noise and three real-world datasets with pre-existing missing values (two biomedical datasets and one time-series dataset). For the former, we show that our proposed GINN method is on par or outperforms several existing state-of-the-art approaches, especially when we consider high levels of injected artificial noise, e.g., up to $50\%$ of missing values in the original dataset. For the latter, we show that our method is robust to the selection of a downstream classifier, with an accuracy comparable to any other combination of an imputation method and a classifier.}

\subsection*{Organization of the paper}

The rest of the paper is organized as follows. In Section \ref{sec:related_work} we describe the relation of this paper with state-of-the-art methods for MDI (Section \ref{subsec:missing_data_imputation}) and graph neural networks (Section \ref{subsec:graph_neural_networks}). The GCN, which is the building block of our method, is described in Section \ref{sec:graph_convolutional_networks}. Then, our graph imputation neural network (GINN) framework and all its extensions are described in Section \ref{sec:proposed_framework}. After a large experimental evaluation in Section \ref{sec:experimental_evaluation}, we provide some concluding remarks in Section \ref{sec:conclusions}.

\section{Related work}
\label{sec:related_work}

\subsection{Missing data imputation}
\label{subsec:missing_data_imputation}

Algorithms for MDI can be categorized depending on whether they perform univariate or multivariate imputation, and on whether they provide one or multiple imputations for each missing datum \citep{van2018flexible}. In addition, different algorithms can make different theoretical assumptions on whether the data is missing completely at random (MCAR) or not. In this paper we consider multivariate imputation, which is standard in the neural network's literature. In the following we briefly review state-of-the-approaches on this topic, including several algorithms that we will compare to, and discuss their relation with our proposal.

A popular technique for MDI is multiple imputation using chained equations (MICE) \citep{azur2011multiple,white2011multiple,van2018flexible}. MICE iteratively imputes each variable in the dataset by keeping the other variables fixed, repeating this for multiple cycles, each time drawing one or more observations from some predictive distribution on that variable. Although MICE has shown very good performance in some settings, especially in the bio-medical sector, the assumptions beyond MICE (especially the MCAR assumption) might result in biased predictions and subsequently lower accuracy \citep{azur2011multiple}.

In the machine learning community, it was recognized very soon that MDI can be framed as a predictive task, on which variants of standard supervised algorithms can be applied, including k-nearest neighbors (k-NN) \citep{acuna2004treatment}, decision trees \citep{lakshminarayan1996imputation}, support vector techniques \citep{wang2006missing}, and several others. However, these techniques have always achieved mixed performance in practice compared to simpler strategies such as mean imputation \citep{bertsimas2017predictive}. k-NN is limited in making weighted averages of similar feature vectors, while other algorithms are required to build a global model of the dataset to be used for imputation. In this paper we also frame the MDI problem in a predictive context, but our proposed model can leverage both global aspects of the dataset and local similarities between different points.

More recently, there has been a surge of interest in applying deep learning techniques to the problem of MDI. These include multiple imputation with deep denoising autoencoders (MIDA) \citep{gondara2017multiple}, combinations of deep networks with probabilistic mixture models \citep{smieja2018processing}, recurrent neural networks \citep{bengio1996recurrent,che2018recurrent}, or generative models including generative adversarial networks \citep{yoon2018gain} and variational autoencoders \citep{nazabal2018handling}. Generally speaking, these methods are better at capturing complex correlations in the data (and in the missing data process), thanks to their multiple layers of nonlinear computations, but they still require to build a global model from the dataset, while ignoring potentially important contributions from similar points. The method we propose can be seen as an extension both of the MIDA algorithm and of \citet{yoon2018gain}, but we focus on a more recent class of NNs, graph NNs, to capture local dependencies. We briefly survey the literature on this topic next.

\subsection{Graph neural networks}
\label{subsec:graph_neural_networks}

Some of the earliest works on extending NNs to the domain of generic graphs were presented in \citet{gori2005new,scarselli2009graph}, and later reformulated in \citet{li2015gated} in a more recent context. These works were mainly motivated by the analogies between unrolled recurrent neural networks and the diffusion of information across a graph.

Another line of work, upon which we build our proposal, considers instead the extension of convolutional neural networks to graph domains under the general term of geometric deep learning \citep{defferrard2016convolutional,kipf2016semi,bronstein2017geometric} (and \citep{micheli2009neural} for earlier works on a similar context). This is done by exploiting recent ideas in the field of graph signal processing \citep{sandryhaila2013discrete,sardellitti2017graph} to define a more general convolution operator able to work on irregular data structures. In particular, in this work we use the GCN of \citet{kipf2016semi}, that for every layer includes a linear diffusion process across neighbors. Additional interesting lines of research in building GNNs that we briefly mention include earlier works on graph autoencoders \citep{sperduti1994encoding}, graph attention networks \citep{velivckovic2017graph}, non-local NNs \citep{wang2018non}, graph embeddings \citep{zhang2018tree2vector}, and tree/graph echo state networks \citep{gallicchio2010graph,gallicchio2013tree}. An overview of many of these ideas is provided in \citet{battaglia2018relational}. We explore some of the ideas from \citet{battaglia2018relational} in our framework by discussing how to include global information about the dataset in the reconstruction process in Section \ref{subsec:including_global_statistics_from_the_data_set}.

Finally, our work is related to the field of manifold regularization \citep{belkin2005manifold,belkin2006manifold}, a semi-supervised class of methods that exploits similarity information among patterns to enforce a regularization term on the optimization process. We build upon them for the construction of our similarity graph, a necessary step for exploiting the power of GNNs.

\section{Graph convolutional networks}
\label{sec:graph_convolutional_networks}

Because the GCN layer is a fundamental building block of our method, we briefly describe it here before moving on to the proposed framework for MDI. Consider a set of $n$ vertices of a directed graph, whose connectivity is described by a (weighted) adjacency matrix $\mathbf{A} \in \R^{n \times n}$, where $A_{ij}$ is different from $0$ if and only if nodes $i$ and $j$ are connected. Each node $i$ has an associated vector of features $\mathbf{x}_i \in \R^d$, that we collect row-wise in the matrix $\mathbf{X} \in \R^{n \times d}$. We would like to have a generic neural network component able to process simultaneously the features at every node, but also take into consideration their relations, expressed through the adjacency matrix.

One way to extend the idea of convolutional networks to this domain is the so-called graph Fourier transform \citep{bruna2013spectral,sandryhaila2013discrete}. Define the Laplacian matrix of the graph as $\mathbf{L} = \mathbf{D} - \mathbf{A}$, where $\mathbf{D}$ is the diagonal degree matrix with $D_{ii} = \sum_{j=1}^n A_{ij}$. We can perform the eigendecomposition of this matrix as $\mathbf{L} = \mathbf{U}\mathbf{\Lambda}\mathbf{U}^T$, where $\mathbf{U}$ is a matrix collecting column-wise the eigenvectors of $\mathbf{L}$, and $\mathbf{\Lambda}$ is a diagonal matrix with the associated eigenvalues. The equivalent of a classical Fourier transform on a signal can be defined in the graph domain as \citep{sandryhaila2013discrete}:

\begin{equation}
\hat{\mathbf{X}} = \mathbf{U}^T\mathbf{X} \,,
\end{equation}

\noindent and the inverse transform as $\mathbf{X} = \mathbf{U}\hat{\mathbf{X}}$. Using this, a straightforward way to define a convolutional layer on graphs \citep{bruna2013spectral} is to first apply the graph Fourier transform, apply a trainable transformation on the frequency components (associated to the eigenvalues of the Laplacian), and then back-transform using the inverse Fourier transform. While viable, this approach is however costly and impractical in most cases.

Later authors \citep{defferrard2016convolutional,kipf2016semi} have noted that by applying a restricted class of filters to the frequency components (polynomials), it is possible to work directly in the graph domain using polynomials of the Laplacian itself. In particular, \citet{kipf2016semi} proposed the GCN with the use of linear filters, resulting in the following canonical layer:

\begin{equation}
\mathbf{H}_1 = g \left( \mathbf{L} \mathbf{X} \mathbf{\Theta}_1 \right) \,,
\label{eq:gcn_layer}
\end{equation}

\noindent where $\mathbf{\Theta}_1$ is a matrix of adaptable coefficients, and $g(\cdot)$ a generic element-wise activation function, such as the ReLU $g(s) = \max\left(0, s\right)$.\footnote{To avoid some numerical instabilities, it is possible to renormalize the Laplacian to properly bound its eigenvalues, ad done in \citet{kipf2016semi}. More in general, one can substitute the Laplacian with any valid graph shift operator \citep{gama2019convolutional}.} Note that the right-multiplication by $\mathbf{\Theta}_1$ is akin to a classical feedforward layer, while the left-multiplication by $\mathbf{L}$ allows to propagate the information across the immediate neighbors of each node. Multiple layers of this form can then be stacked to obtain a complete graph NN. Importantly, for a generic network with $L$ layers of the form \eqref{eq:gcn_layer}, the output of node $i$ will depend on the outputs of its neighbors up to degree $L$. 

\section{Proposed framework for missing data imputation}
\label{sec:proposed_framework}

In MDI, we are also given a data matrix $\mathbf{X}$, which has the same size and semantic as in the previous section, but in general no graph information associated with it. Some of the values of $\mathbf{X}$, denoted by a binary mask $\mathbf{M} \in \left\{0, 1\right\}^{n \times d}$, are missing and need to be imputed for downstream processing or classification/regression. We assume to have either numerical features, which are properly normalized, or categorical features that are represented with one-hot encoding. For other types of features, e.g., text, a previous embedding step is needed \citep{pennington2014glove}.\footnote{When facing a supervised learning problem, for which there is an additional label (e.g., class) associated to each input $\mathbf{x}_i$, we can easily include the training labels in the imputation process by concatenating them to the input vector.}

\begin{figure}
    \centering
     \includegraphics[width=\columnwidth]{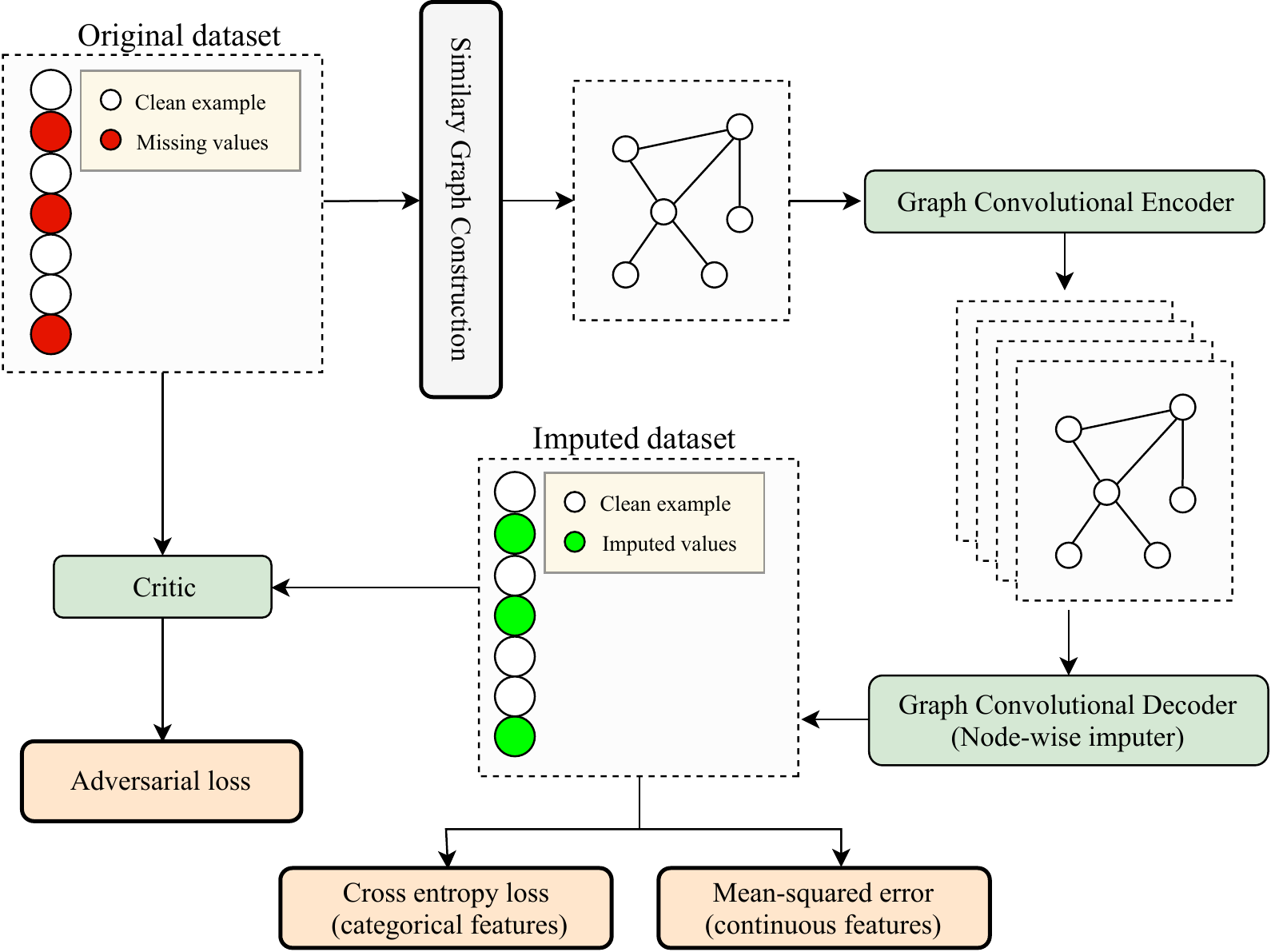}
    \caption{Schematics of the proposed framework for missing data imputation. In green we show the components that are trained end-to-end on the imputation task. In orange we show the different losses. Details on all the steps are provided in the text.}
    \label{fig:framework}
\end{figure}

Predictive models for MDI described previously in Section \ref{subsec:missing_data_imputation} build a function $f(\mathbf{x}_i)$ for imputing missing values of a single example $\mathbf{x}_i$, but in general, do not exploit directly the potentially important information contained in points that might be similar to it. Here, we propose to model this constraint explicitly by building $f$ using GCN blocks, as shown schematically in Figure \ref{fig:framework}. To do this, we first need to build a graph of inter-patterns similarities from $\mathbf{X}$, as described in the next section.

\subsection{Construction of the similarity graph}
\label{sub:graph_construction}
The first fundamental step of our method is the discovery of the graph structure underneath the tabular representation of the data. In the resulting graph, each feature vector in the dataset is encoded as a node of the graph, while the adjacency matrix $\mathbf{A}$ is derived from a similarity matrix $\mathbf{S}$ of the features vectors. As we stated before, constructing a similarity graph from the data is a known problem in the literature, and here we leverage some work from the field of manifold regularization \citep{belkin2005manifold,belkin2006manifold}, adapting it for handling the presence of missing data. A small overview on possible alternatives is provided at the end of this section.

The similarity matrix is computed pairwise for all features vectors using the Euclidean distance, but each time only the non-missing elements of both vectors are used for the computation \citep{eirola2013distance}:

\begin{equation}
S_{ij}  = d(\mathbf{x}_i \odot ( \mathbf{M}_i \odot \mathbf{M}_j), \mathbf{x}_j \odot (\mathbf{M}_i \odot \mathbf{M}_j))
\end{equation}
where $\odot$ stands for the Hadamard product between vectors, $\mathbf{M}_i$ is the $i$th column of the binary matrix $\mathbf{M}$, and $d$ is the Euclidean distance.


\begin{figure}[t]
\centering
\includegraphics[scale=0.4]{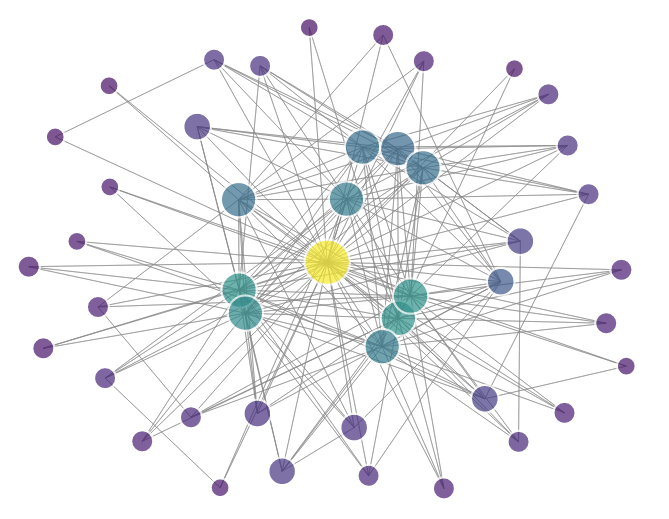}
\caption{Subset of the graph reconstructed from the Iris dataset using our pipeline. For each node the color intensity represents the number of connections  and the size the number of missing features. A clear correlation can be seen between the two.}
\label{fig:graph}
\end{figure}

In practice, it is common to have sparse similarity matrices \citep{belkin2006manifold}, most notably for efficiency and computational reasons (see in particular the discussion on computational cost later on), allowing most operations to scale at-most linearly in the number of neighbors. In order to have a sparse graph, with meaningful connections between similar nodes, we apply a pruning step over the similarity matrix $\mathbf{S}$ inspired from the large-scale manifold learning algorithm in \citet{talwalkar2013large}. A threshold is applied independently over every row of $\mathbf{S}$, by computing a percentile for each row that will act as a threshold, such that only the connections above this threshold inside every row are kept. To make this process more robust, we iterate it twice. The result is used as adjacency matrix for the GCN in the next section.

We found that the $97.72$nd percentile provides good results over all the datasets used in the benchmark of Section \ref{sec:experimental_evaluation}, allowing us to discard at each step around $95\%$ of all the possible connections (in accordance with \citet{talwalkar2013large}). Figure \ref{fig:graph} shows a subset of the graph produced with this procedure starting from the Iris dataset; here we display the relationship between the number of missing features in a node and its degree. It can be seen from the figure how nodes with very few non-zero elements are correlated with a higher number of connections.\footnote{This is due to the absence of a re-normalization step in the computation of the similarity matrix \citep{eirola2013distance}. In practice, we have found this setup to work better than renormalizing all distance measures, possibly because of the increased degree of elements with multiple missing values.}

\textit{On the construction of the similarity graph}: the method described in this section, which is the one we follow in our experiments and in our open-source implementation, was found to provide good empirical performance. Nonetheless, we underline that in the proposed GINN framework, the similarity graph can be built according to any guideline or method available in the literature. For example, classical alternative choices in the semi-supervised literature include binary weights on the edges, heat kernel similarity \citep{belkin2006manifold}, or selecting a fixed number of neighbors instead of a percentile \citep{geng2012ensemble}. If the inputs contain text, images, or similar data, cosine similarity on custom pre-trained embeddings are also a popular choice \citep{bui2018neural}. We leave an analysis of these different alternatives to future work.

\subsection{Autoencoder architecture}
\label{subsec:autoencoder_architecture}

Autoencoders are composed by an encoder which maps the inputs to an intermediate representation in a different dimensional space $\mathbf{h} = \text{encode}(\mathbf{x})$, and a decoder that maps $\mathbf{h} \in \R^m$ to the original dimensional space  $\hat{\mathbf{x}}=\text{decode}(\mathbf{h})$. We use $m > d $ for an overcomplete representation, thus mapping the input into a higher dimensional space with the aim of helping data recovery. Our graph imputer neural network (GINN) will thus be defined as follows:

\begin{equation}
\begin{multlined}
\mathbf{H} = \text{ReLU} \left( \mathbf{L} \mathbf{X} \mathbf{\Theta}_1 \right)
\\\mathbf{\hat X} = \text{Sigmoid}\left( \mathbf{L} \mathbf{H} \mathbf{\Theta}_2 \right)\\
\end{multlined}
\end{equation}

\noindent where $\mathbf{L}$ has been defined in  Section \ref{sec:graph_convolutional_networks} (the extension to networks with multiple hidden layers being straightforward).

Note that we cannot trivially train the autoencoder on the missing values, because they are not known in the training stage. To solve this, we adopt a denoising version of the autoencoder \citep{vincent2008dae}, in which for every optimization step we add additional masking noise on the input, by the means of an inverted dropout layer applied directly on the input of the network. In particular, at each optimization step, we randomly remove $50\%$ of the remaining inputs.\footnote{The only exception: whenever training labels are used as inputs for the imputation process, we do not apply dropout on them.}
In this way, the autoencoder learns to reconstruct any part of the input matrix, similarly to \citep{gondara2017multiple}.

We train the whole model end-to-end minimizing the reconstruction error over the non-missing elements of the dataset. The loss function is thus defined as the combination of a mean squared error (MSE) for the numerical variables and the cross-entropy (CE) for the categorical variables:

\begin{equation}
\label{eq:reconstruction_loss}
L_{A} = \alpha \text{MSE}(\mathbf{X}, \hat{\mathbf{X}}) + (1 - \alpha) \text{CE}(\mathbf{X}, \hat{\mathbf{X}})
\end{equation}
where MSE always returns $0$ for categorical values and \textit{vice versa} for CE. $\alpha$ is an additional hyper-parameter that we initialize as the ratio between the number of numerical columns of the dataset and the total number of columns. Alternatively, it can be tuned like the other hyper-parameters of the network, although we have not found any definite improvement in doing so. 

\subsubsection*{Computational cost of the model}

The computational cost of our approach is related mostly to (a) the one-time cost of constructing the similarity graph, and (b) the use of a GCN layer instead of a standard feedforward layer as in alternative autoencoder architectures \citep{gondara2017multiple}. The cost of point (a) is well-studied in the literature on large-scale similarity search, e.g., \citep{dong2011efficient}, and many techniques and implementations exist for making it efficient. The cost of the GCN layer is discussed in \citet[Section 3.2]{kipf2016semi}. In particular, using a sparse representation for the adjacency matrix reduces the memory requirement to $\mathcal{O}(|E|)$, where $E$ is the number of edges in the graph, and the cost of computing \eqref{eq:gcn_layer} to $\mathcal{O}(|E|CF)$, where $C$ and $F$ are the number of input and output features respectively. In practice, when using an early-stopping strategy for training, we have found the training time for our GINN algorithm (including point (a)) to be significantly faster than alternative neural approaches and on-par with non-neural competitors such as MICE, e.g., see Fig. 4 in the additional materials for a comparison.

\subsection{Adversarial training of the autoencoder}

In order to speed up training, we use an additional adversarial training strategy where a critic, a feedforward network in our case, learns to distinguish between imputed and real data. This is inspired by generative adversarial networks \citep{goodfellow2014} and was found to have significant effects in several reconstructions tasks, particularly in the medical domain \citep{ker2018deep}. In particular, having an adversarial loss during reconstruction forces the reconstructed vector to lie close to the natural distribution of the original patterns \citep{shen2019end}.

To train jointly autoencoder and critic and have a stable training we used the Wasserstein distance introduced in \citet{arjovsky2017}, which is informally defined as the
minimum cost of transporting mass in order to transform a distribution $q$ into a distribution $p$. Using the Kantorovich-Rubinstein duality \citep{villani2008} the objective function is obtained as follows:

\begin{equation}
\min _{A} \max _{C \in \mathcal{D}} \underset{\mathbf{x} \sim \mathbb{P}_{real}}{\mathbb{E}}[C(\mathbf{x})]-\underset{\hat{\mathbf{x}} \sim \mathbb{P}_{imp}}{\mathbb{E}}[C(\hat{\mathbf{x}})) ]
\label{eq:wasserstein_loss}
\end{equation}

\noindent where $\mathcal{D}$ is the set of 1-Lipschitz functions, $\mathbb{P}_{imp}$  is the model distribution implicitly defined by our GCN autoencoder $ \hat{\mathbf{x}}= A(\mathbf{x})$, and  $\mathbb{P}_{real}$ is the unknown data distribution. Practically, Eq. \eqref{eq:wasserstein_loss} can be computed by drawing random mini-batches of data and approximating expectations with averages.


The original loss in \citet{arjovsky2017} used weight clipping to force the Lipschitz property. A further step towards training stability, introduced in \citet{gulrajani2017}, is to use a gradient penalty instead of the weight clipping, obtaining the final loss:

\begin{equation}
   L_{C}=\underset{\hat{\mathbf{x}} \sim \mathbb{P}_{imp}}{\mathbb{E}}[C(\hat{\mathbf{x}})]-\underset{\mathbf{x} \sim \mathbb{P}_{real}}{\mathbb{E}}[C(\mathbf{x})]+\lambda \underset{\tilde{\mathbf{x}} \sim \mathbb{P}_{\tilde{\mathbf{x}}}}{\mathbb{E}}\left[\left(\left\|\nabla_{\tilde{\mathbf{x}}} C(\tilde{\mathbf{x}})\right\|_{2}-1\right)^{2}\right] 
\end{equation}
where $\lambda$ is an additional hyper-parameter. We define $\mathbb{P}_{\tilde{\mathbf{x}}}$ as sampling uniformly from the combination of the real distribution $\mathbb{P}_{imp}$ and from the distribution resulting from the imputation $\mathbb{P}_{imp}$. This means that the feature vector $\tilde{\mathbf{x}}$ will be composed by both real and imputed elements in almost equal quantity.
This  distance is continuous and differentiable thus, the more we train the critic, the more reliable the gradients are. Following standard practice in the GAN literature, in our implementation the GCN autoencoder is trained once for every five optimization steps of the critic, and the total loss becomes:

\begin{equation}
L_D = L_A - \underset{\hat{\mathbf{x}} \sim \mathbb{P}_{imp}}{\mathbb{E}}[C(\hat{\mathbf{x}})]
\end{equation}
since it must fool the critic and minimize the reconstruction error at the same time.


\subsection{Including skip connections in the model}

The autoencoder itself generates an approximate reconstruction of the dataset, while the critic loss guides the autoencoder in this process. However, our main scope is the imputation of values not present in the data. For this task we want a greater contribution coming from the most similar nodes. For this reason, we introduce an additional skip layer which consists always in a graph convolution operation but propagating the information across the immediate neighbors of each node without the node itself. This prevents the autoencoder from learning the identity function.

\begin{figure}[!th]
    \centering
    \includegraphics[width=\columnwidth]{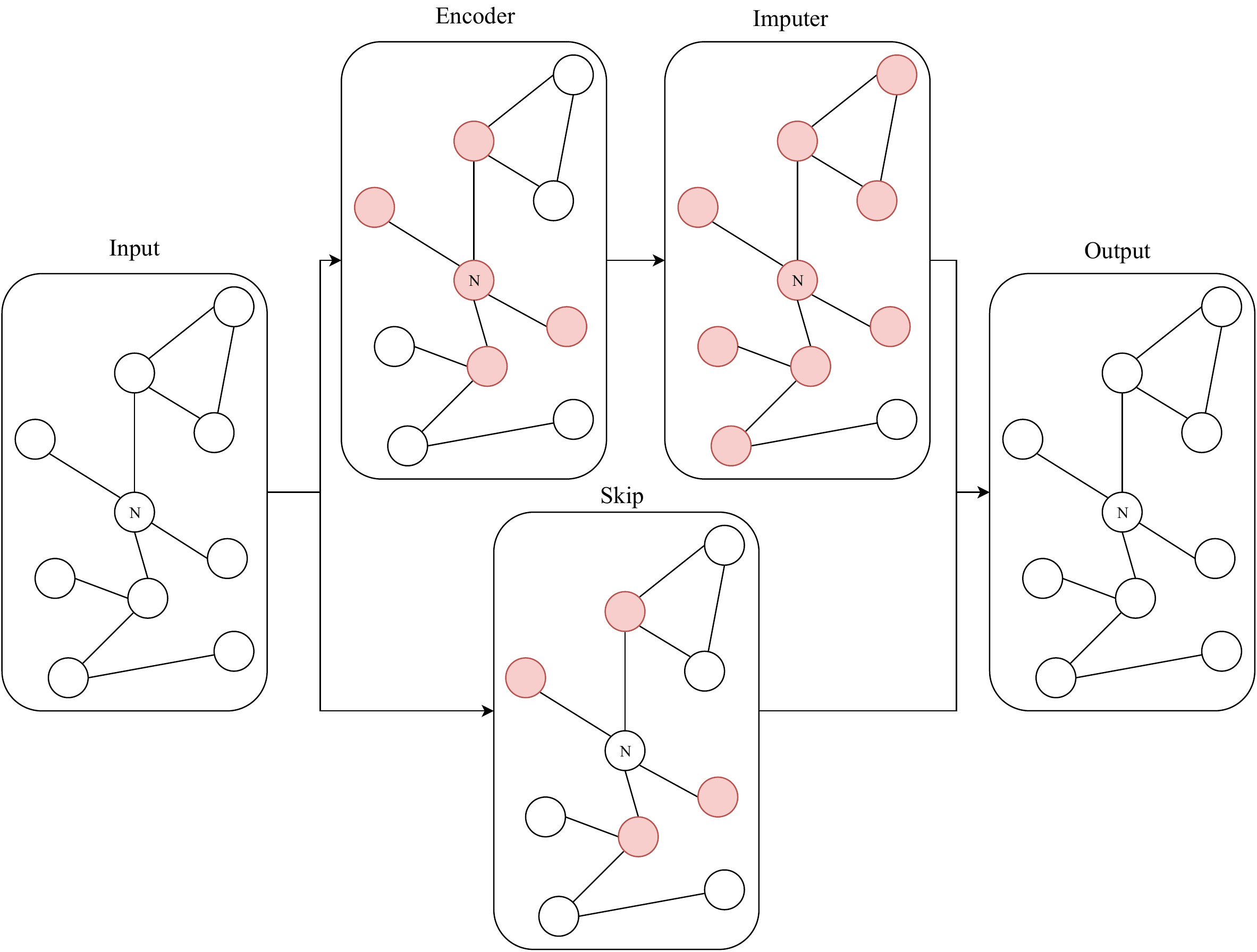}
    \caption{Schematics of the graph autoencoder with the skip connection. We highlight the nodes involved, directly or indirectly by the convolution for the update of the node N. The encoding phase involves 1-hop neighbors, decoding/imputation phase instead involves nodes up to 2-hops. The skip connections involves 1-hop neighbors without considering node N.
    }
    \label{fig:skip_connections}
\end{figure}

The decoding layer becomes:

\begin{equation}
\mathbf{\hat X} = \text{Sigmoid}\left( \mathbf{L} \mathbf{H} \mathbf{\Theta}_2 +  \tilde{\mathbf{L}} \mathbf{X} \mathbf{\Theta}_3 \right )
\end{equation}
where $\tilde{\mathbf{L}}$ is computed similarly to $\mathbf{L}$ (as described in Section \ref{sec:graph_convolutional_networks}), but starting from an adjacency matrix without self loops. This is shown schematically in Figure \ref{fig:skip_connections}.

\subsection{Including global statistics from the dataset}
\label{subsec:including_global_statistics_from_the_data_set}

Another extension we explore is the possibility of including global information on the dataset during the computation of the autoencoder. The inclusion of a global set of attributes, in the context of graph neural networks, was described in-depth by \citet{battaglia2018relational}. 

As a proof of concept, in our case we set as global attribute vector $\mathbf{g}$ for the graph some statistical information of the dataset, including mean or mode of every attribute. The global component can be taken into account in the last layer, weighting their contribution for the update of each node:

\begin{equation}
\mathbf{\hat X} = \text{Sigmoid}\left( \mathbf{L} \mathbf{H} \mathbf{\Theta}_2 +  \tilde{\mathbf{L}} \mathbf{X} \mathbf{\Theta}_3  + \mathbf{\Theta}_4\mathbf{g} \right )
\end{equation}

In addition, if the computation of the global information is differentiable, we can compute a loss term with respect to the global attributes of the original dataset:
\begin{equation}
\begin{multlined}
L = L_d + \gamma \text{MSE}(\text{global}(\mathbf{\hat{X}}), \text{global}(\mathbf{X}))
\end{multlined}
\end{equation}

\noindent where $\gamma$ is some additional weighting term.

\section{Experimental evaluation}
\label{sec:experimental_evaluation}

We divide our experimental evaluation in five subsections. Firstly, following common literature, we evaluate the proposed GINN framework on 20 real-world datasets from the UCI Machine Learning Repository \citep{dua2019} to which we artificially add some desired level of missing values, to evaluate the imputation performances. The characteristics of these 20 datasets are summarized in Table \ref{tab:data}. This selection contains categorical, numerical, and mixed datasets, ranging from 150 observation to 30000 and from only 4 attributes to almost 40. Every dataset is divided into training 70\% and test 30\% sets. Missingness is introduced completely at random on the training set with 4 different levels of noise: 10\%, 20\%, 30\%, and 50\%. Our evaluation will focus first on imputation performance as described in Section \ref{sec:impperf}, then on the accuracy of post-imputation prediction in Section \ref{sec:predperf}. We then perform a comprehensive ablation study of the architecture in Section \ref{sec:ablation}, and an evaluation of the performance of the algorithm on new data in Section \ref{sec:imputation_over_unseen_data}.

Secondly, in Section \ref{sec:evaluation_preexisting_missing_values_new} we evaluate the performance of the algorithm on three real-world datasets with pre-existing (i.e., non artificially induced) missing values. In this part we also evaluate the computational cost of the method when compared to other state-of-the-art approaches. In one case, missing values are also present in the training labels, in which case we provide an application of our method to a semi-supervised scenario (as described later on).

For all the benchmarks, we use an embedding dimension of the hidden layer of $128$, sufficient for an overcomplete representation for all the datasets involved, and we train the model for a maximum of $10000$ iterations with an early stopping strategy for the reconstruction loss over the known elements. The critic used is a simple 3-layer feed-forward network trained 5 times for each optimization step of the autoencoder. We used the Adam optimizer \citep{kingma2014adam} for both networks with a learning rate of $1\times10^{-3}$ and $1\times10^{-5}$ respectively for autoencoder and critic. When label information is available for the datasets, we consider the training labels as an additional feature of each input vector, but we remove this information when processing new (validation or test) data. All experiments are repeated five times and we collect average performance and standard deviation.

All the code for replicating our experiments and using the GINN algorithm is released as an open-source library on the web.\footnote{\url{https://github.com/spindro/GINN}}

\begin{table*}[!th]
\resizebox{\textwidth}{!}{%
\begin{tabular}{l|c|c|c|}
Name & observations & numerical attr.    & categorical attr.\\
\midrule
abalone                    & 4177  & 8  & 0  \\
anuran-calls               & 7195  & 22 & 3  \\
balance-scale              & 625   & 4  & 0  \\
breast-cancer-diagnostic   & 569   & 30 & 0  \\
car-evaluation             & 1728  & 0  & 6  \\
default-credit-card        & 30000 & 13 & 10 \\
electrical-grid-stability  & 10000 & 14 & 0  \\
heart                      & 303   & 8  & 5  \\
ionosphere                 & 351   & 34 & 0  \\
iris                       & 150   & 5  & 0  \\
page-blocks                & 5473  & 10 & 0  \\
phishing                   & 1353  & 0  & 9  \\
satellite                  & 6435  & 36 & 0  \\
tic-tac-toe                & 958   & 0  & 9  \\
turkiye-student-evaluation & 5820  & 0  & 32 \\
wine-quality-red           & 1599  & 11 & 0  \\
wine-quality-white         & 4898  & 11 & 0  \\
wine                       & 178   & 13 & 0  \\
wireless-localization      & 2000  & 7  & 0  \\
yeast                      & 1484  & 8  & 0 
\end{tabular}%
}
\caption{Datasets used for the benchmark. All of them were downloaded from the UCI repository.}
\label{tab:data}
\end{table*}

\subsection{Imputation Performance}
\label{sec:impperf}

This evaluation focuses on the comparison of MAE and RMSE between GINN and 6 other state-of-the-art imputation algorithms: MICE \citep{buuren2011mice}, MIDA \citep{gondara2017multiple}, MissForest \citep{steckhoven2012missforest}, mean \citep{little1986mean}, matrix factorization and k-NN imputation \citep{troyanskaya2001knnmf}. Concerning MICE and MissForest we used the default parameters discussed in the corresponding papers. For the other methods, we fine-tuned the hyper-parameters according to the corresponding literature to provide a fair comparison. In particular, we used $1\times10^{-3}$ and $1\times10^{-4}$ as learning rate for matrix factorization and MIDA with the latter being a 2-layer with 128 units per layer network like our embedding dimension. Finally, we let $k=5$ for the k-NN. The imputation accuracy for each dataset is presented in Table \ref{tab:rmse} for the scenario in which 30\% of the entries are missing, while the results for all other levels of missingness (both in terms of RMSE and MAE) are presented in the supplementary material. We can see from Table \ref{tab:rmse} that the proposed GINN method obtains the best imputation performance in almost half of the datasets, being the second-best in almost all the remaining ones.

\begin{figure}[!th]
    \centering
    \subfigure[Comparison over MAE]{\includegraphics[scale=0.48]{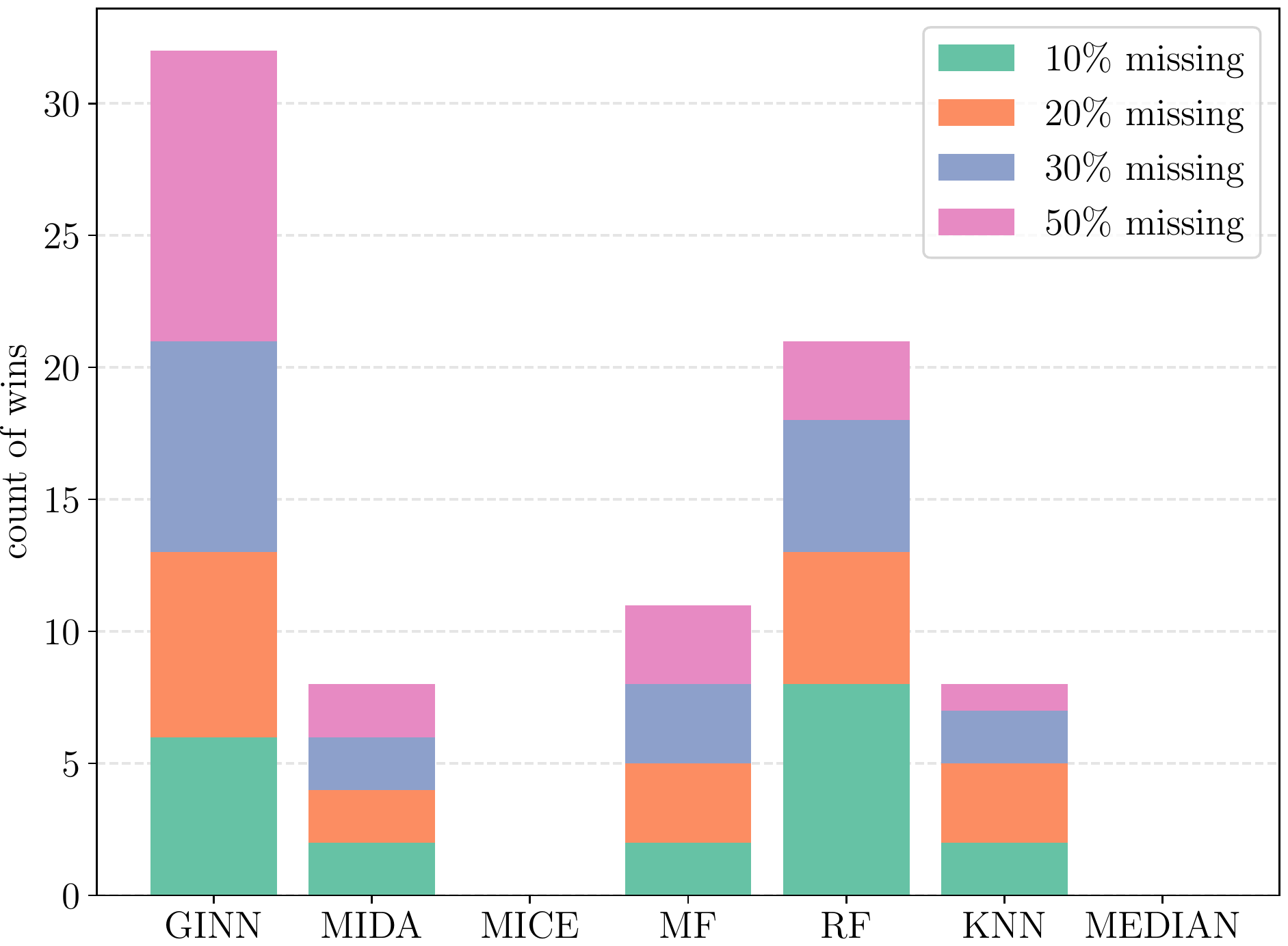}\label{fig:impressub1}}
    \subfigure[Comparison over RMSE]{\includegraphics[scale=0.48]{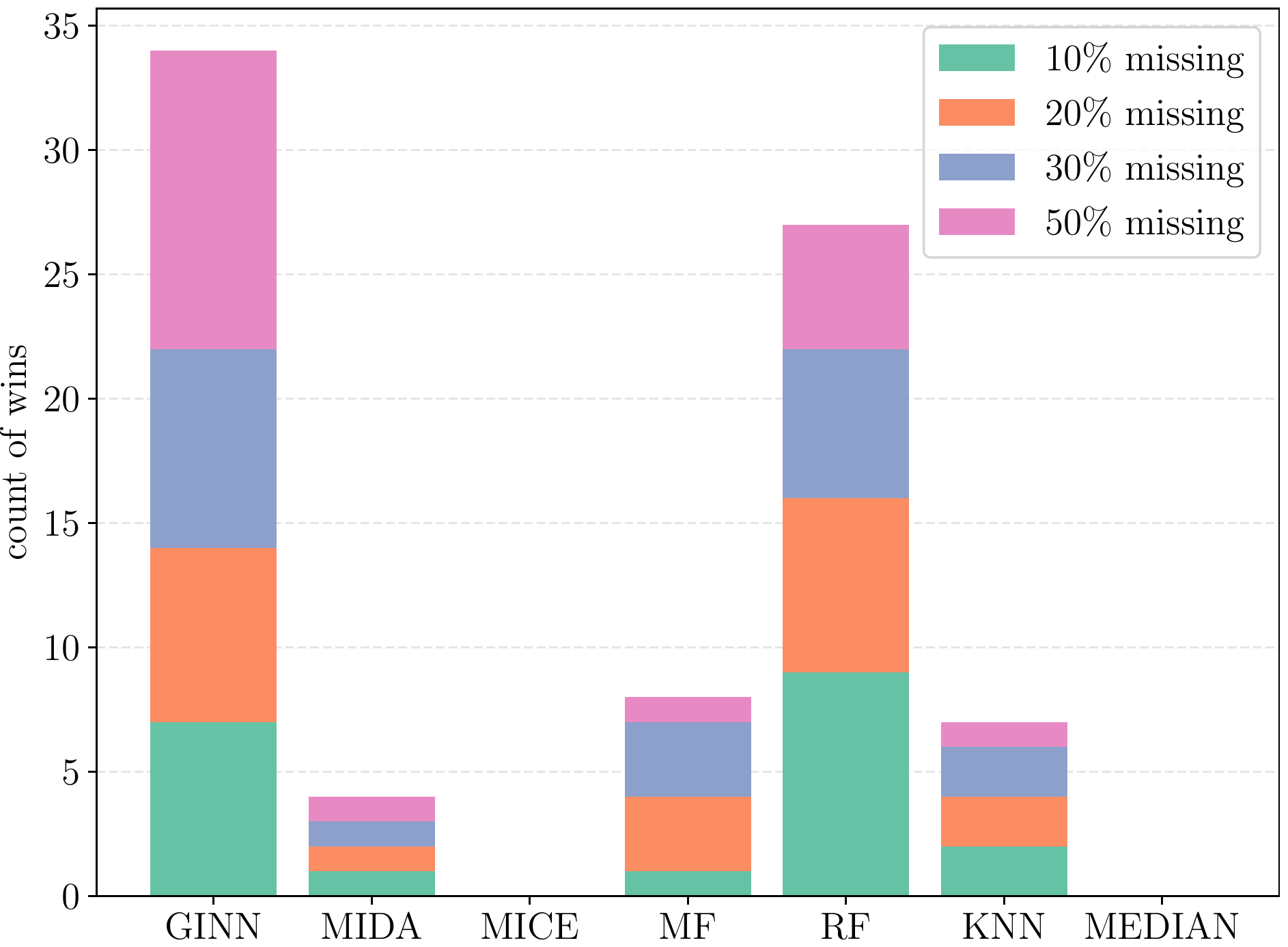}\label{fig:impressub2}}
    \caption{Number of datasets in which each MDI method achieves (a) lowest average MAE or (b) lowest average RMSE from the true values. The different colors of the bar stand for different percentages of missing elements: from the bottom 10\% at the lowest to the top 50\% at the highest.}
    \label{fig:impres}
\end{figure}

To provide a more schematic comparison, in Figure \ref{fig:impres} we show the summary of those results for every level of missing data. In these histograms, we provide the number of times that each method achieves the best imputation (on average), i.e. the lowest RMSE in Figure \ref{fig:impressub1} and MAE in Figure \ref{fig:impressub2}. Concerning the lower percentage of missing elements (10\%, 20\%) our method is almost on par with the best among the algorithms tested, i.e., MissForest. When those percentages increase our method brings a huge improvement over the state-of-the-art with the highest difference at 50\% where our method significantly outperforms all other techniques. Aggregating the results obtained at 30\% and 50\% percentage of missing features, we have that our method is the best in 50\% of the cases against the 27.5\% of its best competitor MissForest, when looking at the MAE, and 47.5\% against 20\% for the RMSE. We defer a statistical analysis of these results to the next subsection, where we analyze also the results for a downstream predictive task.

\begin{sidewaystable}
\resizebox{\textwidth}{!}{%
\begin{tabular}{l|c|c|c|c|c|c|c|}
 RMSE & GINN    & MIDA    & MICE&  MF    & RF&     k-NN&     MEDIAN\\
\midrule
abalone  &  0.904  $\pm$ 0.048  &  1.190  $\pm$ 0.002  &  1.069  $\pm$ 0.053  &  3.321  $\pm$ 2.220  &  \textbf{0.811}  $\pm$ 0.008  &  1.072  $\pm$ 0.035  &  1.112  $\pm$ 0.001 \\
anuran-calls  &  0.070  $\pm$ 0.009  &  0.187  $\pm$ 0.001  &  0.076  $\pm$ 0.000  &  0.091  $\pm$ 0.000  &  0.154  $\pm$ 0.003  &  \textbf{0.058} $\pm$  0.000  &  0.226  $\pm$ 0.002 \\
balance-scale &  0.559  $\pm$ 0.010  &  \textbf{0.436}  $\pm$ 0.001   &  0.579  $\pm$ 0.005  &  0.521  $\pm$ 0.021  &  0.516  $\pm$ 0.000 &  0.580  $\pm$ 0.002  &  0.577  $\pm$ 0.001  \\
breast-cancer-diagnostic  &  54.633  $\pm$ 4.519  &  114.390  $\pm$ 5.612  &  38.122  $\pm$ 3.568  &  23.559  $\pm$ 2.964  &  \textbf{21.155}  $\pm$ 2.401  &  39.653  $\pm$ 6.124  &  127.645  $\pm$ 6.907  \\
car-evaluation  &  \textbf{0.623}  $\pm$ 0.003  &  0.632   $\pm$ 0.003  &  0.636  $\pm$ 0.003  &  0.845  $\pm$  0.000  &  0.636  $\pm$ 0.002 &  0.636  $\pm$ 0.006  &  0.649  $\pm$ 0.002 \\
default-credit-card  &  17479.409  $\pm$ 280.991  &  19664.115  $\pm$ 197.618  &  15687.056   $\pm$ 262.499  &  15212.927  $\pm$ 388.170  &  \textbf{12897.406}  $\pm$ 279.296  &  17294.412  $\pm$ 56.008  &  23708.027  $\pm$ 100.798   \\
electrical-grid-stability   &  1.534  $\pm$  0.005 &  1.857  $\pm$ 0.003   &  1.637  $\pm$ 0.005   &  \textbf{1.474} $\pm$ 0.023  &  1.536  $\pm$ 0.008  &  1.763  $\pm$ 0.028  &  1.555  $\pm$ 0.001  \\
heart  &  \textbf{10.883}  $\pm$ 2.003   &  25.212  $\pm$ 1.926  &  12.326  $\pm$ 1.506  &  11.367  $\pm$ 1.745  &  11.368   $\pm$ 1.814 &  14.846 $\pm$ 1.560   &  11.708  $\pm$ 1.726   \\
ionosphere   &  0.425  $\pm$ 0.001   &  1.118  $\pm$ 0.005   &  0.462  $\pm$ 0.007   &  340.257  $\pm$ 20.027  &  0.407  $\pm$ 0.006  &  \textbf{0.380} $\pm$ 0.001  &  0.551   $\pm$   0.004 \\
iris  &  0.349  $\pm$ 0.047   &  1.792  $\pm$ 0.106  &  0.379  $\pm$ 0.003  &  \textbf{0.106}  $\pm$ 0.269   &  0.333   $\pm$ 0.008   &  0.446  $\pm$ 0.062   &  1.159 $\pm$ 0.053  \\
page-blocks  &  \textbf{536.645}  $\pm$ 44.594  &  1035.750  $\pm$ 80.323  &  1889.068  $\pm$ 6.830  &  795.441  $\pm$ 85.233  &  579.227   $\pm$  147.154  &  604.171  $\pm$ 33.633   &  869.476  $\pm$ 49.517   \\
phishing   &  \textbf{0.493}  $\pm$ 0.006   &  0.529  $\pm$ 0.006  &  0.625  $\pm$ 0.000  &  0.594  $\pm$ 0.000  &  0.606  $\pm$ 0.020  &  0.548  $\pm$ 0.001 &  0.581  $\pm$ 0.002  \\
satellite  &  6.256   $\pm$ 0.200   &  11.528  $\pm$ 0.112    &  4.481  $\pm$ 0.013   &  4.562  $\pm$ 0.024     &  \textbf{3.634}  $\pm$ 0.005    &  4.523  $\pm$ 0.040    &  18.335  $\pm$ 0.060   \\
tic-tac-toe  &  \textbf{0.469}  $\pm$ 0.010    &  0.576   $\pm$ 0.000    &  0.657   $\pm$  0.001   &  0.816   $\pm$  0.005   &  0.671   $\pm$  0.005   &  0.625    $\pm$ 0.001    &  0.622   $\pm$ 0.001  \\
turkiye-student-evaluation  &  0.295   $\pm$ 0.002    &  \textbf{0.285}   $\pm$ 0.000    &  0.537    $\pm$ 0.001    &  0.287   $\pm$  0.000   &  0.292   $\pm$  0.001   &  0.303    $\pm$0.000      &  0.515  $\pm$ 0.001  \\
wine   &  \textbf{47.901}    $\pm$ 7.298    &  154.344    $\pm$  22.959   &  52.358   $\pm$ 14.401    &  54.764  $\pm$  6.209    &  48.672   $\pm$ 8.529    &  54.636   $\pm$ 8.874   &  99.651   $\pm$ 10.909  \\
wine-quality-red  &  \textbf{8.099}     $\pm$ 0.097    &  10.313   $\pm$ 0.091    &  9.092    $\pm$ 0.079    &  8.883   $\pm$ 0.100     &  8.414   $\pm$  0.071   &  9.752   $\pm$  0.079   &  10.265  $\pm$ 0.031  \\
wine-quality-white  &  11.188   $\pm$ 1.133  &  15.383  $\pm$ 0.351    &  11.174   $\pm$ 0.183     &  41.324    $\pm$ 0.196    &  \textbf{10.350}    $\pm$ 0.153    &  12.454   $\pm$  0.391   &  13.432  $\pm$ 0.223  \\
wireless-localization  &  4.851    $\pm$ 0.227    &  8.470   $\pm$ 0.131    &  4.728   $\pm$  0.047   &  \textbf{1.863}   $\pm$ 0.113     &  4.214    $\pm$ 0.105    &  4.995   $\pm$ 0.010    &  8.588  $\pm$ 0.129   \\
yeast   &  \textbf{0.085}     $\pm$   0.000   &  0.162    $\pm$ 0.002    &  0.090   $\pm$ 0.000    &  0.093   $\pm$  0.000   &  0.086    $\pm$ 0.000    &  0.091   $\pm$  0.001   &  0.102   $\pm$ 0.002  
\end{tabular}%
}
\caption{Average Root Mean Squared Error with 30\% of missing elements. Best results for each dataset is highlighted with a bold font. One standard deviation is also provided.}
\label{tab:rmse}
\end{sidewaystable}

\subsection{Predictive Performance}
\label{sec:predperf}
Now we evaluate the performance of standard machine learning algorithms for classification, both binary and multi-class, trained on the various imputations analyzed previously. We consider 4 different classifiers: a k-NN classifier with $k=5$, regularized logistic regression, C-Support Vector Classification with an RBF kernel and a random forest classifier with 10 estimators and a maximum depth of 5. All hyper-parameters are initialized with their default values in the scikit-learn implementation.\footnote{\url{https://scikit-learn.org/stable/modules/classes.html}}

\begin{figure}[!th]
    \centering
    \subfigure[k-NN]{\includegraphics[scale=0.35]{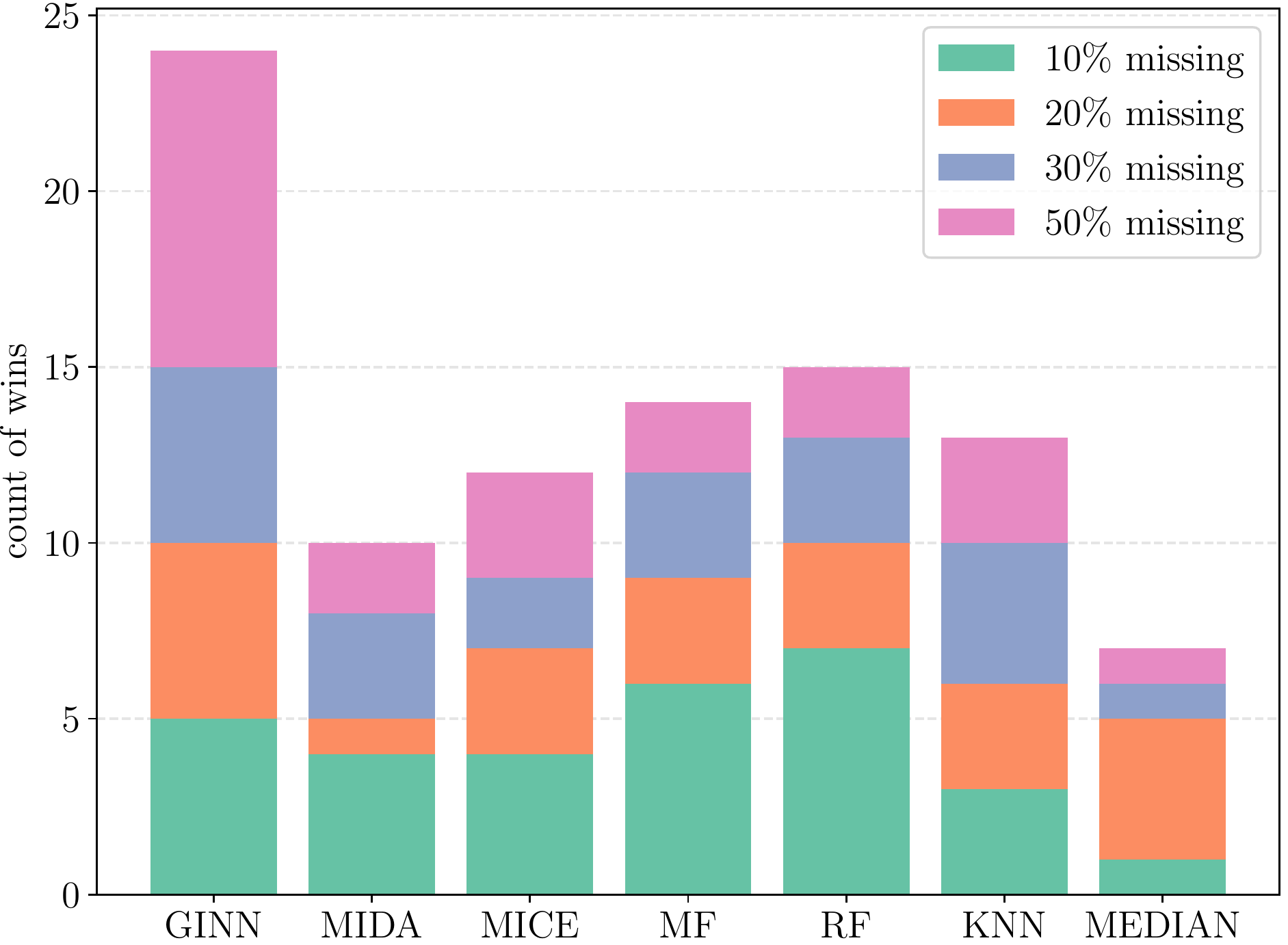}}
    \subfigure[Logistic regression]{\includegraphics[scale=0.35]{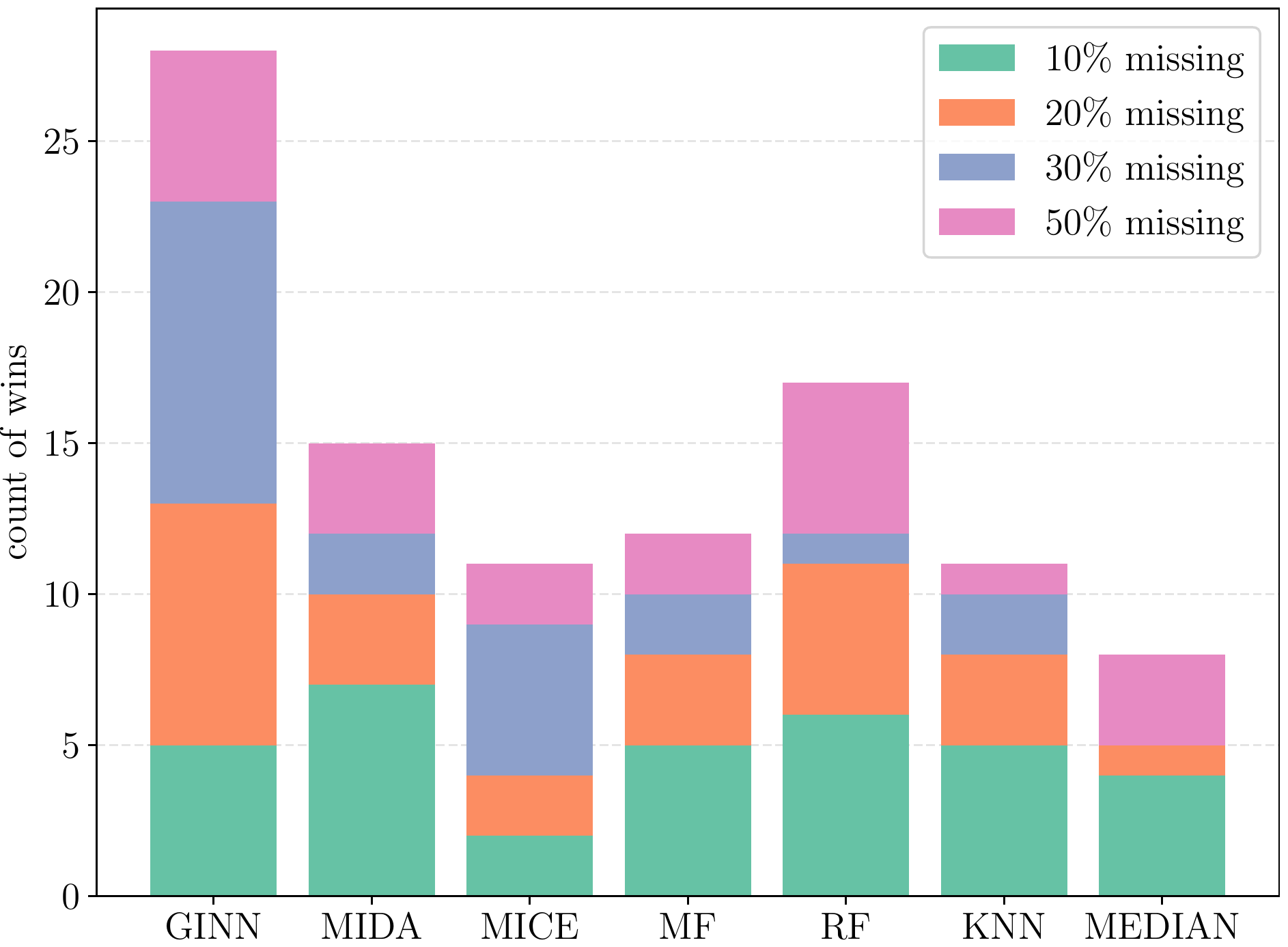}}
    \subfigure[Random forest]{\includegraphics[scale=0.35]{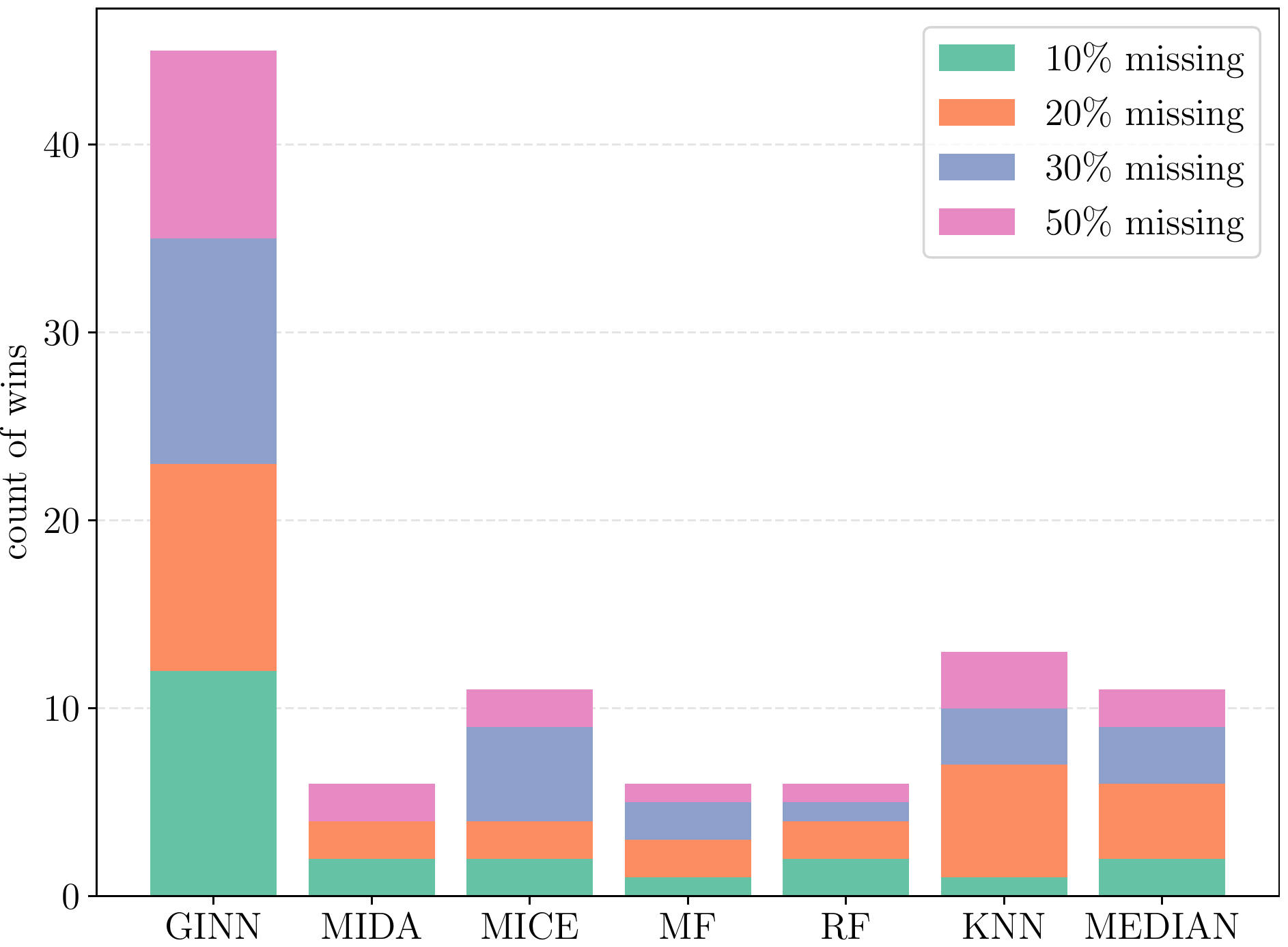}}
    \subfigure[SVC]{\includegraphics[scale=0.35]{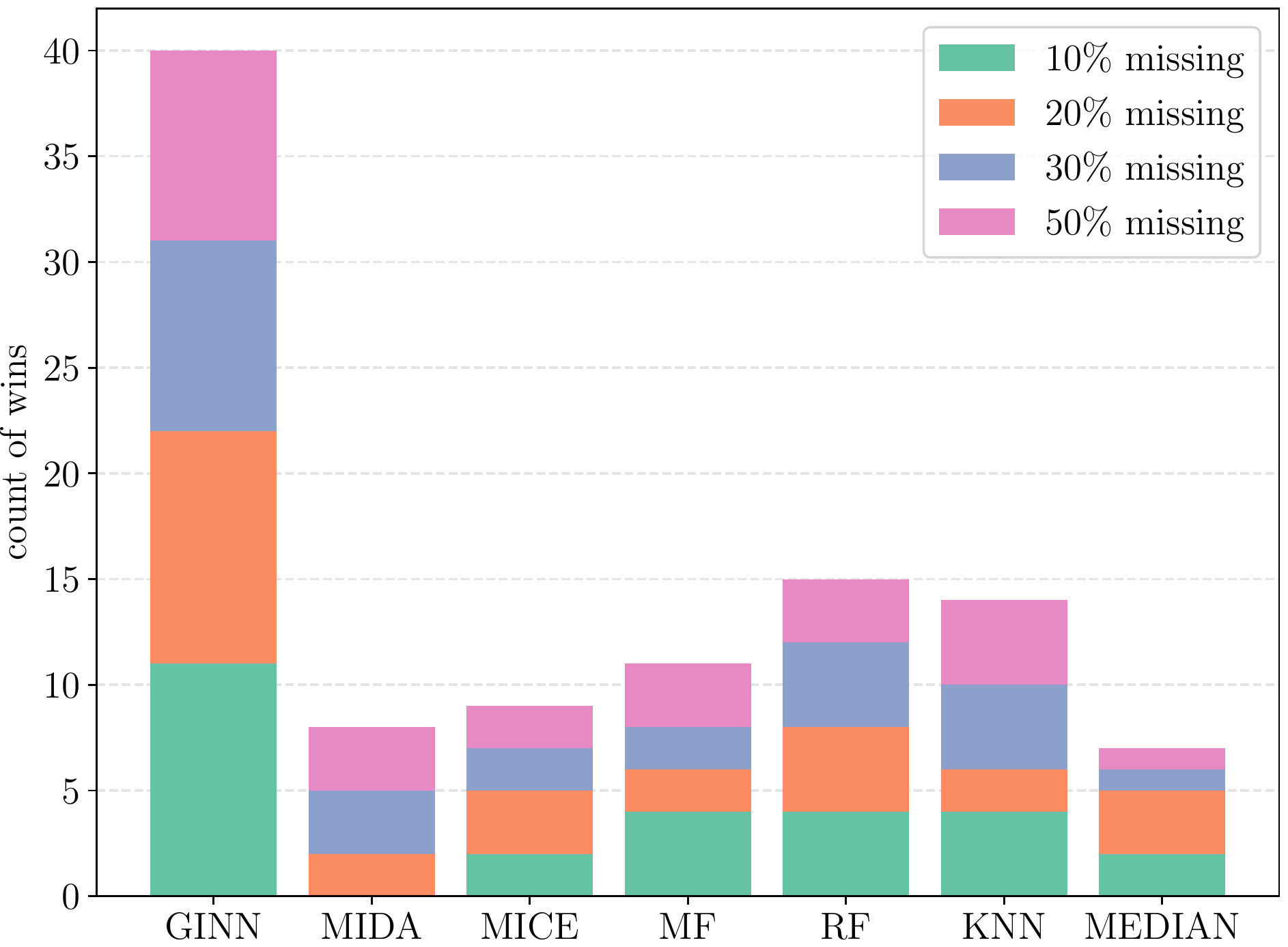}}
    \caption{Number of datasets in which each MDI method achieves the highest classification accuracy for each classifier used. The different colors of the bar stand for different percentages of missing elements: from the bottom 10\% the lowest to the top 50\% the highest.}
    \label{fig:predres}
\end{figure}

The classification accuracy is presented in Table \ref{tab:class} for the scenario in which 30\% of the entries in the data matrix are missing, assuming MCAR, with a random forest classifier. In Figure \ref{fig:predres} we show the summary of the results for each noise level and for each classifier. In these histograms we analyze the number of times each imputation technique allows the classifier to achieve the best average accuracy. In this comparison, we consider also the draws. 

Our method outperforms competitors with every classification algorithm tested, and it has the highest number of wins, winning in $85.62\%$ of the cases. Our method worked very well when paired with SVC and Random forest, improving the accuracy for every percentage of missing elements. With the logistic regression and k-NN classifiers, at the lowest missing percentage (10\%), our method is slightly below the state-of-the-art. As missing percentages increase, we have a huge improvement over the other competitors. This reflects in the findings of the previous Section and confirms the ability of our method of being very successful in the case of moderate to severely damaged datasets.

\begin{sidewaystable}
\resizebox{\textwidth}{!}{%
\begin{tabular}{l|c|c|c|c|c|c|c|c|c|}
 Random forest  & baseline & & GINN & MIDA & MICE & MF & RF & k-NN & MEDIAN \\
\midrule
abalone  &  53.269   $\pm$ 0.804  &  &  \textbf{55.821}  $\pm$ 1.569   &  53.269  $\pm$ 1.169   &  54.944  $\pm$ 1.037  &  51.754  $\pm$ 1.1675   &  52.472  $\pm$ 1.2036  &  53.030  $\pm$ 0.040  &  51.674  $\pm$ 1.555 \\
anuran-calls &  92.728  $\pm$ 0.463  &  &  \textbf{94.441}  $\pm$ 0.787  &  89.810  $\pm$ 1.760  &  91.848  $\pm$ 0.648  &  93.191  $\pm$ 0.556  &  92.635  $\pm$ 0.394   &  91.060  $\pm$ 0.162  &  90.412 $\pm$ 1.135  \\
balance-scale &  81.328  $\pm$ 0.0687  &  &  76.595  $\pm$ 0.452  &  72.340  $\pm$ 0.982  &  76.595  $\pm$ 1.053  &  77.659  $\pm$ 1.301  &  76.063  $\pm$ 0.834  &  \textbf{78.723}  $\pm$ 0.893 &  69.148 $\pm$ 1.140 \\
breast-cancer-diagnostic   &  97.076  $\pm$ 0.877  &  &  96.491 $\pm$0.585  &  95.906  $\pm$ 2.636  &  94.736  $\pm$ 0.877  &  94.152  $\pm$ 1.170  &  97.076  $\pm$ 0.877  &  96.491  $\pm$ 1.170  &  \textbf{97.660}  $\pm$ 1.170  \\
car-evaluation &  70.520  $\pm$ 0.674  &  &  \textbf{72.832}  $\pm$ 0.586  &   71.483  $\pm$ 0.096  &  69.942  $\pm$ 0.096  &  69.942  $\pm$ 0.000  &  70.712  $\pm$ 0.393  &  70.520  $\pm$ 0.482  &  69.942  $\pm$ 0.000  \\
default-credit-card &  77.870  $\pm$ 0.100 &  &  \textbf{77.980} $\pm$ 0.001 &  77.880  $\pm$ 0.000 &  77.880  $\pm$ 0.000   &  77.886  $\pm$ 0.042  &  77.893  $\pm$ 0.017  &  77.9  $\pm$0.025    &  77.9  $\pm$ 0.083  \\
electrical-grid-stability  &  98.533   $\pm$ 0.033 &   &  91.100  $\pm$ 1.633  &  86.634   $\pm$ 0.867  &  97.556   $\pm$ 1.650  &  94.990   $\pm$ 3.967 &  97.867   $\pm$ 0.000 &  98.700   $\pm$ 0.333 &  \textbf{99.360}  $\pm$ 1.233  \\
heart   &  82.417   $\pm$ 2.747  &  &  \textbf{83.516}   $\pm$ 0.549  &  73.626   $\pm$ 2.198 &  \textbf{83.516}  $\pm$ 1.648  &  78.021   $\pm$ 2.198  &  82.417  $\pm$ 1.648 &  82.417  $\pm$ 0.549  &  76.923   $\pm$ 3.297   \\
ionosphere  &  90.566   $\pm$ 0.001     &  &  \textbf{95.283}   $\pm$ 2.358   &  84.905   $\pm$ 1.415  &  92.452   $\pm$ 0.000   &  92.452  $\pm$ 1.415  &  92.452  $\pm$ 0.943  &  \textbf{95.283}   $\pm$ 1.415  &  93.396   $\pm$ 1.415  \\
iris &  91.111   $\pm$ 0.000   &  &  88.889  $\pm$ 1.667  &  80.000   $\pm$ 1.667  &  \textbf{93.334}   $\pm$ 0.000  &  91.112   $\pm$ 3.333   &  91.112  $\pm$ 3.333  &  \textbf{93.334}   $\pm$ 1.667  &  86.667  $\pm$ 0.000  \\
page-blocks &  94.640   $\pm$ 0.594  &  &  \textbf{95.066}   $\pm$ 0.137  &  94.823   $\pm$  0.000 &  94.762   $\pm$ 0.137  &  94.640 $\pm$ 0.046  &  94.640   $\pm$ 0.046  &  94.884   $\pm$ 0.137  &  94.701  $\pm$ 0.000  \\
phishing &  84.729   $\pm$ 1.107   &  &  \textbf{84.482}   $\pm$ 0.738  &  83.004   $\pm$ 0.738   &  83.251  $\pm$ 0.369  &  83.990   $\pm$ 0.923  &  83.004   $\pm$ 0.738  &  84.236  $\pm$ 0.554  &  82.019  $\pm$ 0.185 \\
satellite  &  83.990   $\pm$ 0.207  &  &  \textbf{83.850}   $\pm$ 1.157  &  81.699   $\pm$ 0.155  &  82.550   $\pm$ 0.181  &  83.650 $\pm$ 0.958  &  83.300  $\pm$ 0.233  &  83.500  $\pm$ 0.363  &  79.950  $\pm$ 0.984  \\
tic-tac-toe &  69.443   $\pm$ 0.868  &  &  72.569  $\pm$ 0.174  &  68.056   $\pm$ 0.694  &  67.708  $\pm$ 0.868 &  67.361   $\pm$ 1.042  &  65.972  $\pm$ 4.688  &  71.180  $\pm$ 1.910  &  \textbf{72.916}  $\pm$ 0.174  \\
turkiye-student-evaluation &  84.020   $\pm$ 0.200  &  &  \textbf{84.650}   $\pm$ 0.859  &  84.593   $\pm$ 0.143  &  81.214  $\pm$ 0.601  &  \textbf{84.650}   $\pm$ 0.830  &  83.276   $\pm$ 0.086  &  83.505  $\pm$ 1.088  &  83.218  $\pm$ 0.286  \\
wine  &  96.296   $\pm$ 0.000  &  &  \textbf{98.148}   $\pm$ 0.000 &  92.592   $\pm$ 1.852  &  \textbf{98.148}  $\pm$ 0.926  &  94.444  $\pm$ 0.926  &  \textbf{98.148}   $\pm$ 0.000  &  94.444  $\pm$ 0.926  &  94.444  $\pm$ 3.704 \\
wine-quality-red  &  63.334   $\pm$ 1.655  &  &  62.708  $\pm$ 2.105   &  62.083   $\pm$ 3.635  &  61.667  $\pm$ 2.105  &  \textbf{63.334}   $\pm$ 1.766   &  62.291  $\pm$ 1.042  &  61.458  $\pm$ 1.766  &  58.125  $\pm$ 1.655 \\
wine-quality-white  &  52.108   $\pm$ 0.544  &  &  \textbf{53.673}   $\pm$ 0.952  &  51.360   $\pm$ 0.578  &  52.517  $\pm$ 0.374  &  52.312  $\pm$ 1.224  &  53.265  $\pm$0.204  &  51.020  $\pm$ 0.136  &  51.836  $\pm$ 0.204 \\
wireless-localization  &  97.166   $\pm$ 0.167  &  &  96.334  $\pm$ 1.083  &  93.167   $\pm$ 1.417  &  \textbf{97.334}  $\pm$ 0.833  &  96.667  $\pm$ 0.333  &  97.000   $\pm$ 0.167  &  95.834  $\pm$ 0.083  &  97.167  $\pm$ 0.083 \\
yeast  &  55.156   $\pm$ 0.561  &  &  56.502  $\pm$ 1.457  &  46.188   $\pm$ 3.027  &  \textbf{58.071}   $\pm$ 0.785  &  43.721  $\pm$ 0.897  &  54.484  $\pm$ 0.112  &  56.502  $\pm$ 2.018  &  53.363  $\pm$ 2.018   
\end{tabular}%
}

\caption{Classification accuracy on each dataset using a Random forest classifier. The model was trained over the imputed data by algorithms for 30\% MCAR. In the first column (baseline) we have the accuracy obtained utilizing the undamaged dataset.}
\label{tab:class}
\end{sidewaystable}

We corroborate the results of this and the previous section by performing a statistical analysis of the algorithms according to the guidelines in \citet{demvsar2006statistical}. A Friedman rank test confirms that there are statistical significant differences both with respect to the RMSE of Figure \ref{fig:impres} (p-value of $1.11e^{-11}$), and with respect to the accuracy of the classifiers in Figure \ref{fig:predres} (e.g., p-value for random forest is $1.01e^{-10}$). A successive set of Nemenyi post-hoc tests further confirms statistical significant differences between GINN and all other methods for the random forest classification in Figure \ref{fig:predres}, and between GINN and all other methods except RF for the imputation results in Figure \ref{fig:impres}. The full set of rankings and of p-values for these tests can be found in the additional material for this paper.

\subsection{Ablation study}
\label{sec:ablation}

To investigate how much each step described in Section \ref{sec:proposed_framework} improves our method, we supervised the quality of imputation and convergence by starting from a basic autoencoder. As the starting point we used a 2-layer denoising autoencoder (DAE), obtained by setting the adjacency matrix to be the identity, obtaining a method similar to \citep{gondara2017multiple}. Then we introduced the graph and the graph convolutional layer in our imputer (GINN) followed by the addition of the critic and the adversarial training (A-GINN), the skip connection (A-GINN skip) and finally the global attributes (A-GINN skip global).

\begin{figure}
    \centering
    \subfigure[]{\includegraphics[scale=0.5]{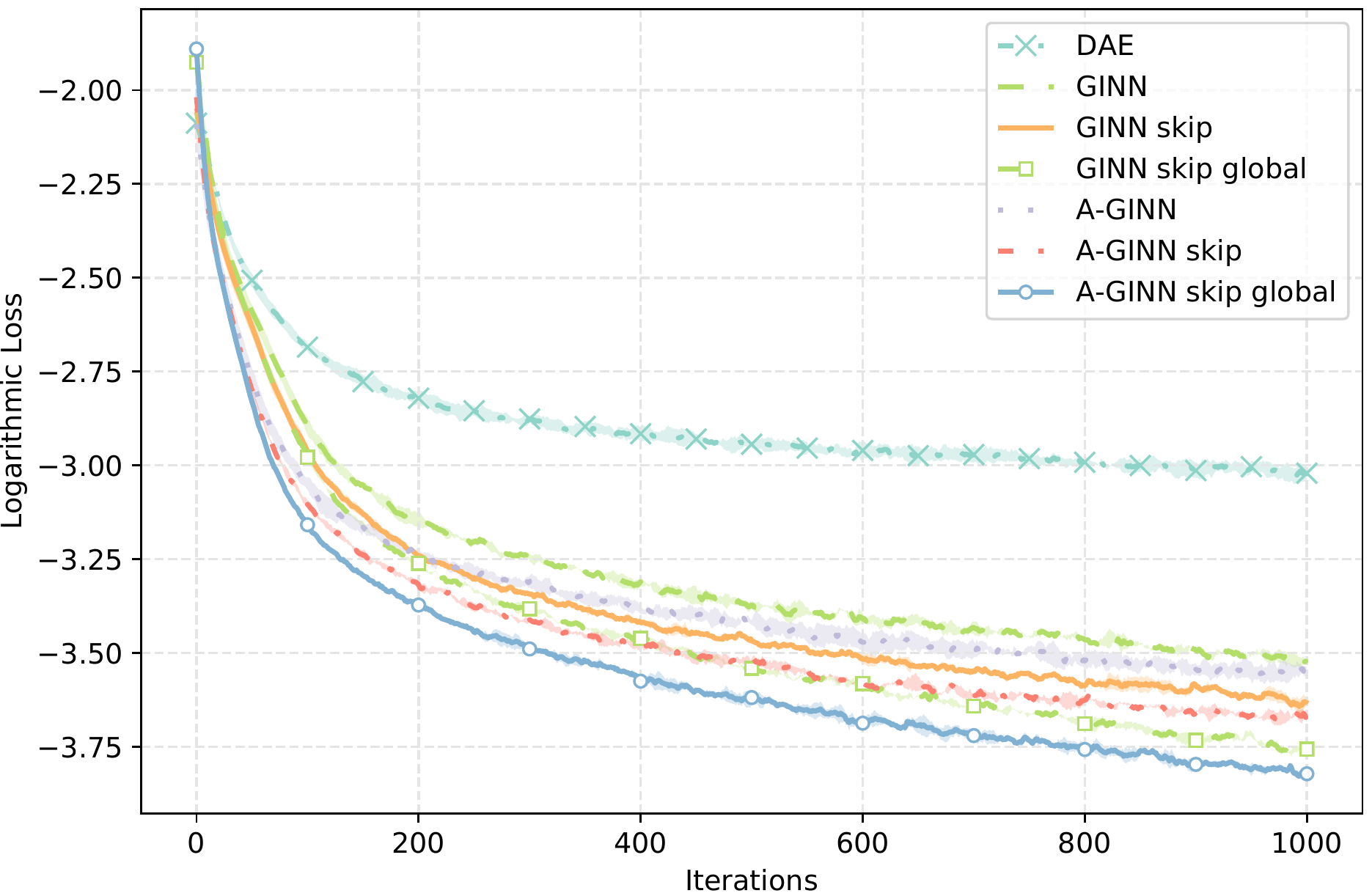}}
    \caption{Convergence of the logarithmic loss function defined in Equation \ref{eq:reconstruction_loss}, on the Ionosphere dataset for each improvement described in this section and with imputation results shown in Table \ref{tab:ablation}.}
    \label{fig:ablation_convergence}
\end{figure}

Regarding imputation performances, Table \ref{tab:ablation} shows how the introduction of the graph and the graph-convolution operation over three randomly selected datasets makes a huge difference against a standard autoencoder. After that, each following step helps to refine even further the imputation accuracy. The reconstruction loss in Eq. \eqref{eq:reconstruction_loss}, more precisely its logarithm, shows a similar behaviour, shown in Figure \ref{fig:ablation_convergence}. After each step we have a better convergence. Similar results are obtained for all the other datasets.


\begin{table*}[t]
\centering
\begin{tabular}{l|c|c|c|}
  & Ionosphere & Tic-Tac-Toe & Phishing \\
\midrule

DAE & $0.309 \pm 0.011$ & $0.323 \pm0.004$ & $0.260 \pm0.005$ \\
GINN & $0.263 \pm0.013$ & $0.317 \pm0.003$  & $0.247 \pm0.002$ \\
GINN skip & $0.258 \pm0.013$ & $0.314 \pm0.002$  & $0.246 \pm0.007$ \\
GINN skip global& $0.256 \pm0.013$ & $0.313 \pm0.000$  & $0.245 \pm0.007$ \\
A-GINN &  $0.257 \pm0.011$ & $0.316 \pm0.001$ &  $0.243 \pm0.006$ \\
A-GINN skip & $0.256 \pm0.013$ & $0.305 \pm0.007$ & $0.243 \pm0.006$ \\
A-GINN skip global & $\textbf{0.255} \pm0.016$  & $\textbf{0.303} \pm0.003$ & $ \textbf{0.241} \pm0.004$ \\
\end{tabular}
\caption{Mean Absolute Error, $\pm$ the standard deviation, of the imputation over 3 trials with  20\% of missing elements. Each dataset has the corresponding convergence shown in Figure \ref{fig:ablation_convergence}.}
\label{tab:ablation}
\end{table*}

The results in Figure \ref{fig:ablation_convergence} are shown with respect to the number of iterations. In the supplementary material, we also provide a similar analysis with respect to a fixed computational budget, while an analysis of the overall computational time when compared to the other algorithms is provided later on in Section \ref{sec:evaluation_preexisting_missing_values_new}.

\subsection{Imputation over unseen data}
\label{sec:imputation_over_unseen_data}

In this section we test the ability of the model of imputing a new damaged portion of the dataset that was not available at training time.
In order to impute these new values, we have first to inject the new data in the graph, adding nodes and edges. We compute a new similarity matrix for the new features (not considering the labels) and also the similarity between these new features and the older ones. Then we add the new nodes to the graph and the edges resulting from the double threshold procedure described in Section \ref{sub:graph_construction}. 

To evaluate this, we introduced missing values and evaluated the imputation performances with and without fine-tuning the model on a second randomly kept portion of the datasets. In Table \ref{tab:new} we show the MAE of the imputation over this new portion of dataset and compare against the other MDI algorithms. We used the same dataset and settings of the ablation study. The fine-tuned version consists of an additional 500 epochs of training over the new graph. It can be seen how our method is able to perform a state-of-the-art imputation on new unseen data without performing additional training and how it improves in case of a small number of additional optimization steps.

\begin{table*}[t]
\centering
\begin{tabular}{l|c|c|c|}
MAE & Ionosphere & Tic-Tac-Toe & Phishing \\
\midrule
GINN   & $ \underline{0.252} \pm 0.008 $ & $ \underline{0.320} \pm 0.012  $ & $ \underline{0.242} \pm 0.002$  \\
GINN (FT)  & $ \textbf{0.235}  \pm 0.004 $ & $ \textbf{0.299} \pm 0.005 $ & $ \textbf{0.237} \pm 0.002$ \\
MIDA   & $ 0.782 \pm 0.000 $ & $ 0.390 \pm 0.004$ & $ 0.422 \pm 0.006$\\
MICE   & $ 0.335 \pm 0.008$ & $ 0.414 \pm 0.004 $ & $ 0.644 \pm 0.003 $\\
MF  & $ 0.312 \pm 0.009$ & $ 0.667 \pm 0.003$ & $ 0.462 \pm 0.029 $\\
RF   & $ 0.254 \pm 0.024$ & $ 0.456 \pm 0.001 $ & $ 0.344 \pm 0.000$\\
k-NN   & $ \textbf{0.235} \pm 0.002 $ & $ 0.400 \pm 0.006$ & $ 0.282 \pm 0.005 $ \\
MEDIAN   & $ 0.362 \pm 0.004$ & $ 0.388 \pm 0.004$ & $ 0.343 \pm  0.008$
\end{tabular}
\caption{Mean Absolute Error of the imputation with 20\% of missing elements.}
\label{tab:new}
\end{table*}

\subsection{ Evaluation on datasets with pre-existing missing values}
\label{sec:evaluation_preexisting_missing_values_new}

To evaluate GINN  in a real-world scenario, we compared its performance against the other state-of-the-art algorithms over three datasets with \textit{pre-existing} missing values. When the data is not missing completely at random, the problem gets more complex, because there may exist relationships between the probability of a variable to be missing and other observed data. We consider two biomedical datasets: mammographic mass introduced in \citet{elter2007mammo} and cervical cancer by \citet{fernandes2017cncr} respectively with 4\% and 13\% of missing elements. We performed the imputation without considering the additional information of the label. Then we solved the binary classification task as done in Section \ref{sec:predperf}.
The third dataset is a time-series of air quality measurements \citep{devito2010air} with 13\% of missing elements. Differently from the other two, missing values can also be found in the labels, making this a semi-supervised task. We select the three initial most damaged months as training data and perform imputation on all data, including the target variable. Then, we performed the downstream task of classifying, in the two successive months, the target variable discretized in three bins. {\color{bostonuniversityred}A summary of the three datasets is provided in Table \ref{tab:real_data}.}

\begin{table}
	\color{bostonuniversityred}
	\linespread{1.0} 
	\setlength{\tabcolsep}{3pt} 
	\renewcommand{\arraystretch}{0.5} 
	\centering
	\taburulecolor{bostonuniversityred}
	\begin{tabu}{l|c|c|c|c}
		Name & observations & numerical attr. & categorical attr. & missingness\\
		\midrule
		c. cancer                & 860  & 9  & 25 & 13\% \\
		m. mass              & 960  & 0 & 5 & 4\%\\
		air quality                    & 3313  & 9  & 0 & 13\%
	\end{tabu}
	\caption{Datasets with pre-existing missingness used for the benchmark in Section 5.5. All of them were downloaded from the UCI repository.}
	\label{tab:real_data}
\end{table}

{\color{bostonuniversityred}
	In Table \ref{tab:real} we show the average accuracy (computed over 10 trials) for every combination of imputation method and downstream classifier. For clarity, we highlight in bold the best result, and we underline the second-best one. While our proposed algorithm does not necessarily achieve the best accuracy overall, it can be seen from  Table \ref{tab:real} that it is, in average, the most resilient to the choice of an external classifier. Overall, when combined with a logistic regression we obtain the best accuracy for the cervical cancer dataset (on par with several other algorithms), while we obtain the second-best result when combined with a random forest or an SVC in the other two cases. In order to highlight the resilience of the algorithm, in the supplementary material we provide a statistical analysis when the results from the different classifiers are aggregated.
	
	Concerning computational performance, we provide execution times for all the algorithms (for simplicity, in the random forest case) in the supplementary material. Briefly, our algorithm is faster than alternative neural approaches (as already described in Section \ref{sec:ablation}), but slower than non-neural approaches such as k-NN.
}

\begin{table}
	\color{bostonuniversityred}
	\linespread{1.0} 
	\setlength{\tabcolsep}{3pt} 
	\renewcommand{\arraystretch}{0.5} 
	\centering
	\taburulecolor{bostonuniversityred}
	\begin{tabu}{l|p{20mm}|p{27mm}|p{25mm}|p{29mm}}
		& Classifier & M. Mass & C. Cancer & A. Quality \\
		\midrule
		GINN & k-NN \newline LR \newline RF \newline SVC & 82.45$\pm$0.8 \newline 82.81$\pm$1.6 \newline 83.04$\pm$1.2 \newline \underline{83.98$\pm$1.3} & 98.35$\pm$0.6 \newline \textbf{99.44$\pm$0.4} \newline 98.82$\pm$0.5 \newline 97.71$\pm$0.1 & 90.09$\pm$0.6 \newline 90.72$\pm$0.1 \newline \underline{90.86$\pm$0.5} \newline 90.08$\pm$0.0 \\
		\midrule
		MIDA & k-NN \newline LR \newline RF \newline SVC & 81.18$\pm$0.7 \newline 81.70$\pm$1.5 \newline 78.50$\pm$2.8 \newline 83.67$\pm$1.4 & 98.00$\pm$0.6 \newline \underline{98.97$\pm$0.6} \newline 97.77$\pm$0.2 \newline 97.71$\pm$0.1 & 88.29$\pm$0.6 \newline 89.53$\pm$0.1 \newline 88.99$\pm$0.9 \newline 88.49$\pm$0.0 \\
		\midrule
		MICE & k-NN \newline LR \newline RF \newline SVC & 80.45$\pm$1.2 \newline 82.41$\pm$1.2 \newline 80.19$\pm$1.9 \newline 83.75$\pm$1.1 & 97.83$\pm$0.4 \newline \textbf{99.44$\pm$0.4} \newline 97.83$\pm$0.4 \newline 97.71$\pm$0.1 & 80.99$\pm$0.8 \newline 81.58$\pm$0.0 \newline 78.14$\pm$0.9 \newline 78.56$\pm$0.1 \\
		\midrule
		MF & k-NN \newline LR \newline RF \newline SVC & 80.57$\pm$1.2 \newline 82.27$\pm$1.1 \newline 77.66$\pm$3.6 \newline \textbf{84.03$\pm$1.2} & 98.45$\pm$0.7 \newline \textbf{99.44$\pm$0.4} \newline 97.78$\pm$0.2 \newline 97.71$\pm$0.1 & 84.41$\pm$0.5 \newline 84.72$\pm$1.2 \newline 83.78$\pm$1.4 \newline 83.29$\pm$0.0 \\
		\midrule
		RF & k-NN \newline LR \newline RF \newline SVC & 81.02$\pm$1.4 \newline 82.67$\pm$1.1 \newline 76.66$\pm$6.8 \newline 83.71$\pm$1.0 & 97.78$\pm$0.2 \newline \textbf{99.44$\pm$0.4} \newline 97.89$\pm$0.5 \newline 97.71$\pm$0.1 & 88.41$\pm$0.4 \newline 88.30$\pm$1.7 \newline 88.65$\pm$1.1 \newline 88.36$\pm$0.0 \\
		\midrule
		KNN & k-NN \newline LR \newline RF \newline SVC & 81.61$\pm$1.2 \newline 82.37$\pm$1.8 \newline 80.11$\pm$2.0 \newline \textbf{84.03$\pm$1.2} & 98.10$\pm$0.6 \newline \textbf{99.44$\pm$0.4} \newline 97.81$\pm$0.3 \newline 97.71$\pm$0.1 & 89.38$\pm$0.5 \newline \textbf{91.18$\pm$0.0} \newline 89.24$\pm$0.5 \newline 89.79$\pm$0.0 \\
		\midrule
		MEDIAN & k-NN \newline LR \newline RF \newline SVC & 80.37$\pm$1.0 \newline 82.17$\pm$1.1 \newline 76.74$\pm$5.8 \newline \textbf{84.03$\pm$1.2} & 98.06$\pm$0.7 \newline \textbf{99.44$\pm$0.4} \newline 97.98$\pm$0.3 \newline 97.71$\pm$0.1 & 88.85$\pm$0.4 \newline 88.46$\pm$0.0 \newline 89.83$\pm$0.5 \newline 88.36$\pm$0.0
	\end{tabu}
	
	\caption{Classification accuracy and standard deviation over 10 trials obtained by using the different imputation algorithms on datasets having pre-existing missing values. For each dataset, we highlight in bold the best result, and we underline the second-best one.}
	\label{tab:real}
\end{table}

\section{Conclusions and future work}
\label{sec:conclusions}

In this paper we introduced a novel technique for missing data imputation, where we used a novel graph convolutional autoencoder to reconstruct the full dataset. We also describe several improvement to our technique, including the use of an adversarial loss, and the inclusion of global information from the dataset in the reconstruction phase. {\color{bostonuniversityred}We show through an extensive numerical simulation that our method has good imputation performance, and the results are robust to the selection of an additional classifier later on. In experiments with a large level of artificial noise, our method is also shown to significantly outperform competitors.}

Future work can consider the adoption of different graph neural architectures for the autoencoding process (such as those mentioned in Section \ref{sec:related_work}), or the extension to other types of noisy data beyond vector-valued data and different types of similarity measures. {\color{bostonuniversityred}In addition, in order to further improve accuracy and training time, we can think of training our imputation module together with a classification step in a end-to-end fashion.} 

Currently, the major drawbacks of our method are the need for computing the similarity matrix of the data, and the difficulty of performing mini-batching in the presence of graph-based data. Both problems are well-known in the corresponding literature, and in future work we plan on investigating techniques for speeding up similarity search and mini-batching on the graph to improve the computational complexity of the method.

\section*{References}
\bibliographystyle{elsarticle-harv}
\bibliography{biblio}

\clearpage
\appendix
\renewcommand\thefigure{\arabic{figure}}
\renewcommand\thetable{\arabic{table}}

\section*{Additional material}

\subsection*{Detailed RMSE results (Section 5.1)}

In Table \ref{tab:rmse_additional} we provide the individual RMSE imputation values, where each row is one of the $20$ datasets and each column is an imputation method (abbreviations are explained in the main text). We separate the results with respect to the level of artificial corruption of the original dataset: the suffix \textit{$\_$xx} means $xx \%$ of missing values that have been artificially added. These results are aggregated and commented in Fig. 4 and Tab. 3 of the main text. Results for MAE are similar and can be found on our online repository.

\subsection*{Detailed accuracy for regression/classification (Section 5.2)}

In Table \ref{tab:rf} we report the accuracy of the downstream classification/regression task for each choice of classification/regression technique when using a random forest technique (results are similar for other methods, and for brevity we provide them on our online repository). Like before, we separate the results with respect to the level of artificial corruption of the original dataset: the suffix \textit{$\_$xx} means  $xx \%$ of missing values that have been artificially added. These results are aggregated and commented in Fig. 5(c) and Tab. 4 of the main text.

\subsection*{Detailed results for the statistical tests}

\begin{figure}[!th]
    \centering
    \subfigure[Rankings]{\includegraphics[width=0.3\columnwidth]{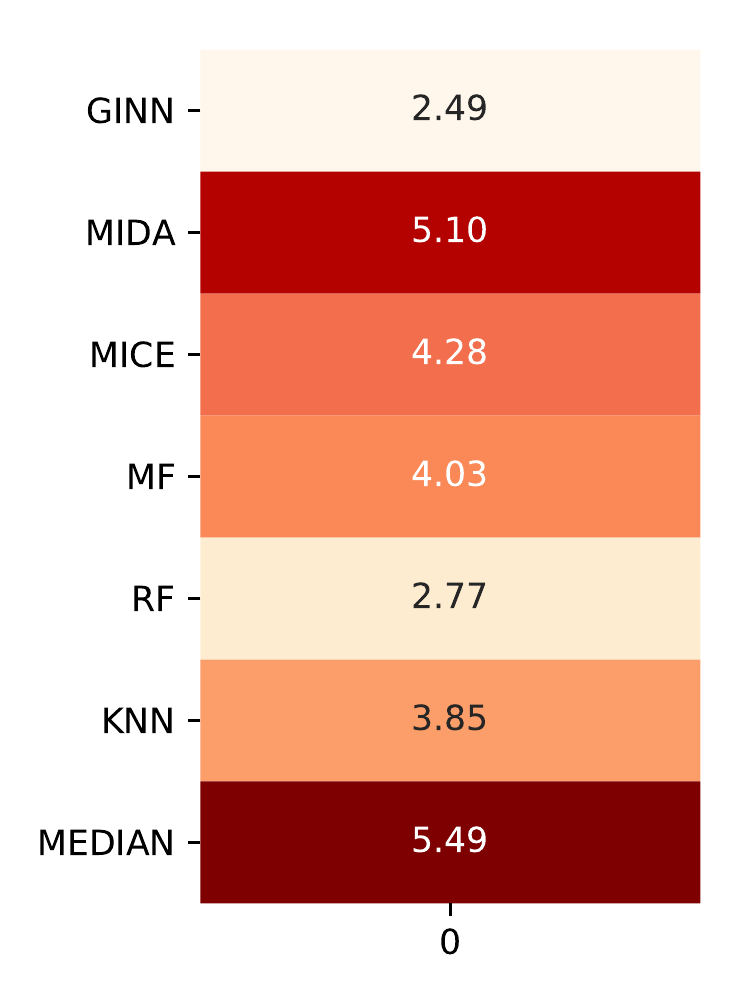}}\
    \subfigure[P-values]{\includegraphics[width=0.8\columnwidth]{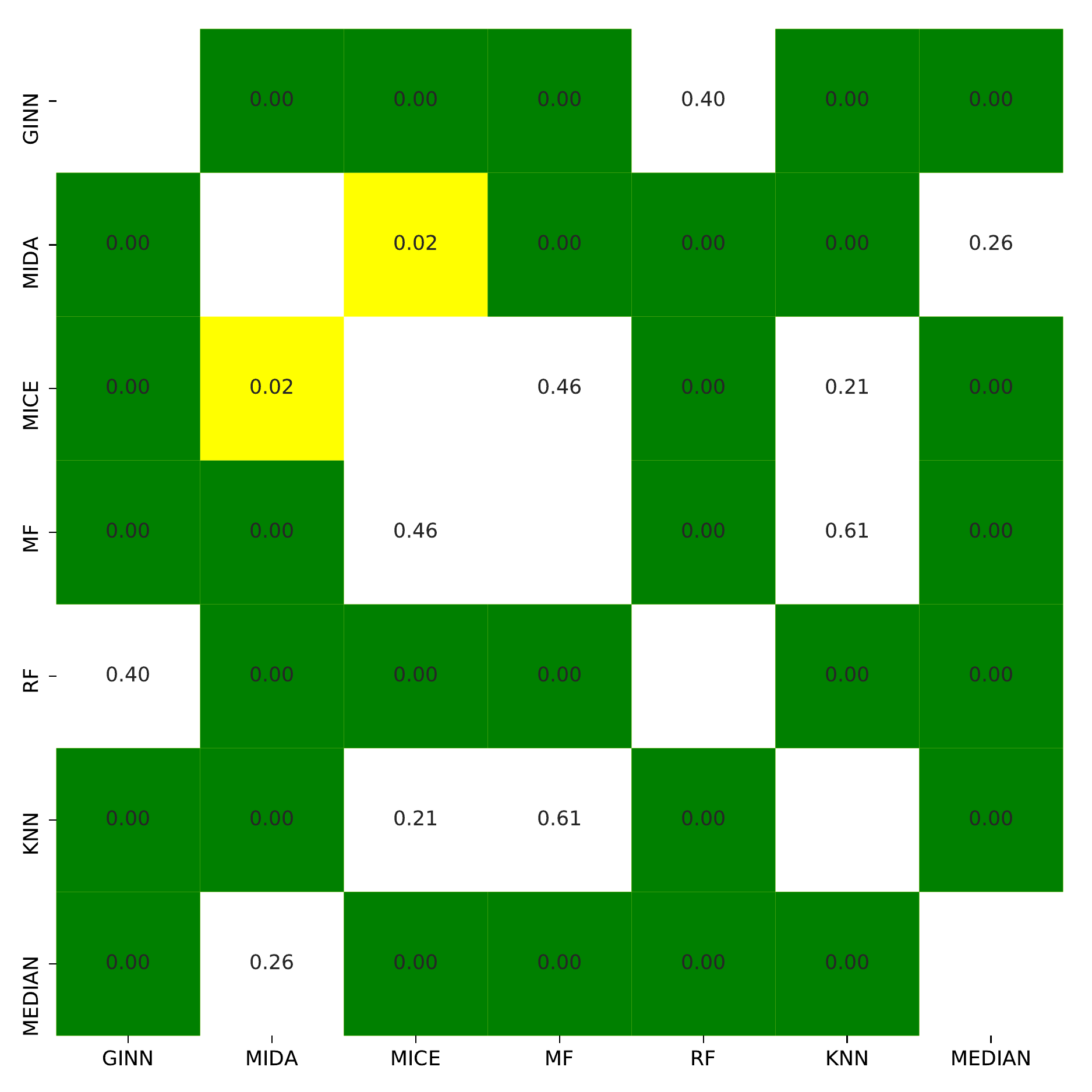}}
    \caption{(a) Rankings of the algorithms in Section 5.1 with respect to RMSE, averaged over all datasets. (b) Corresponding p-values of a set of post-hoc Nemenyi tests. Green shows significant differences at 0.01, yellow significant differences at 0.05.}
    \label{fig:stat_rmse}
\end{figure}

\begin{figure}[!th]
    \centering
    \subfigure[Rankings]{\includegraphics[width=0.3\columnwidth]{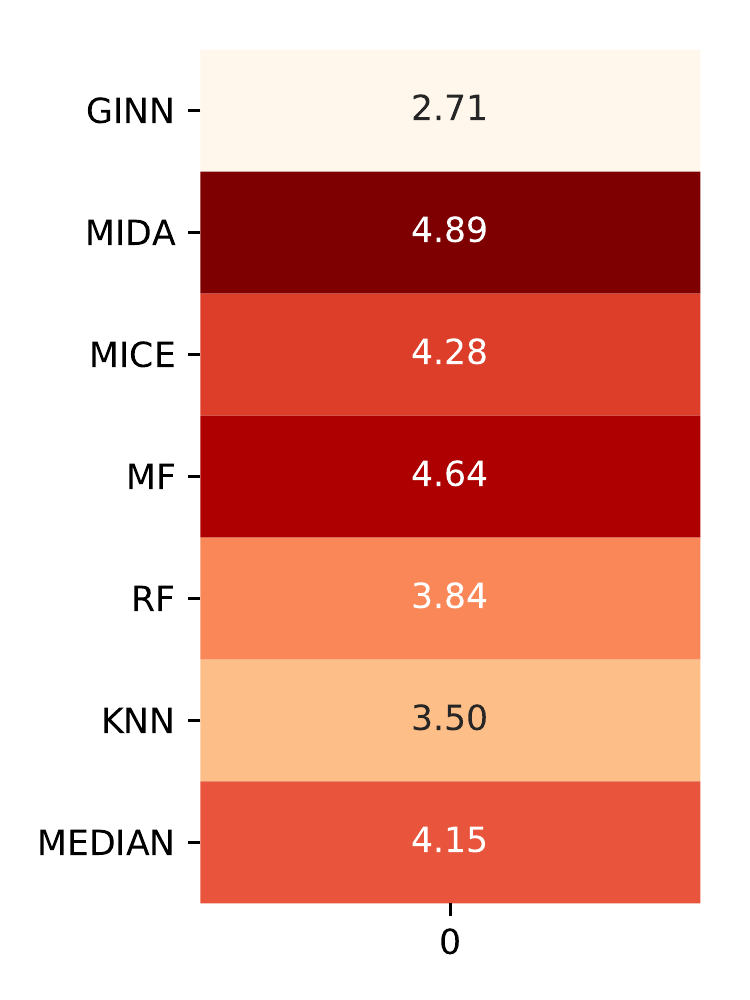}}\
    \subfigure[P-values]{\includegraphics[width=0.8\columnwidth]{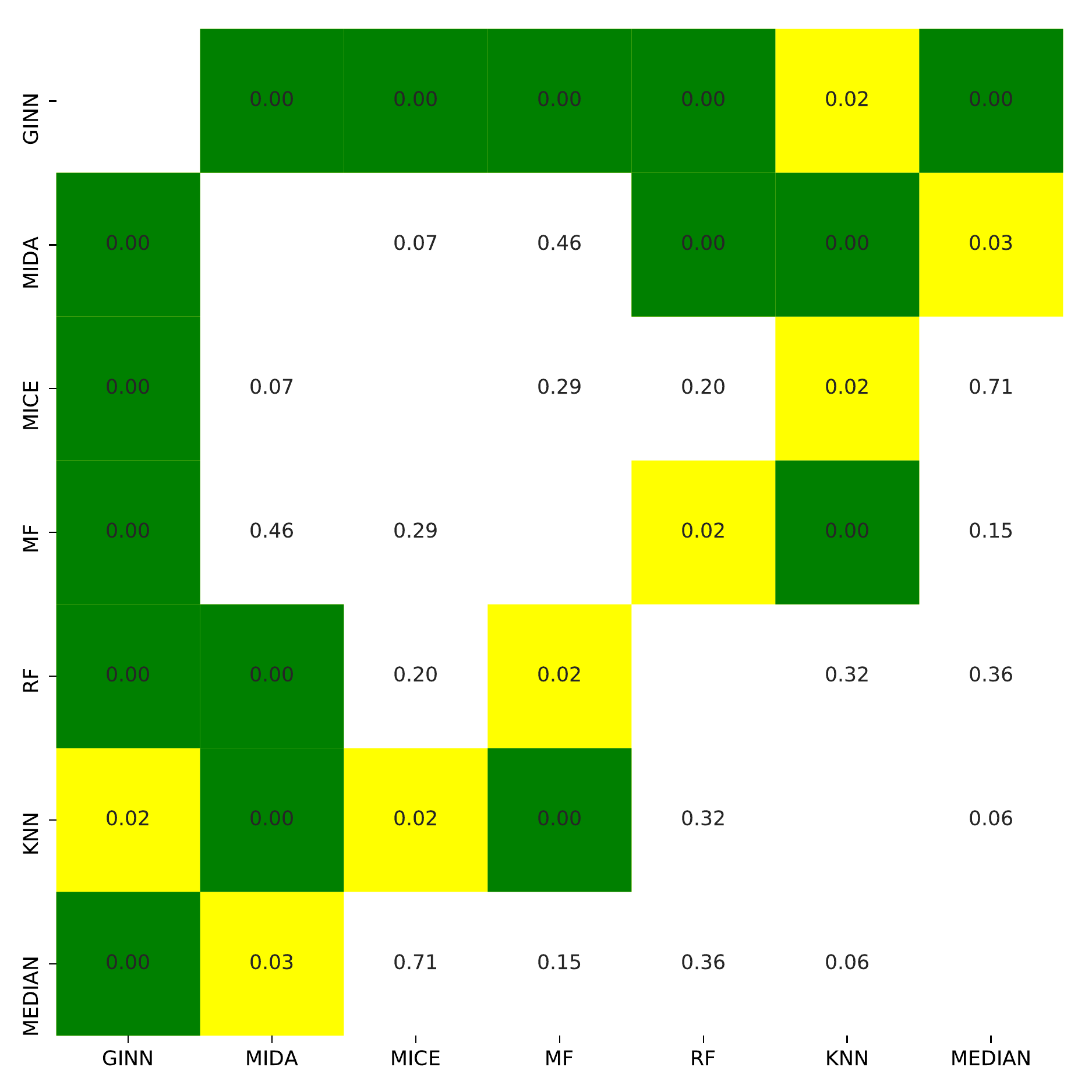}}
    \caption{(a) Rankings of the algorithms in Section 5.2 with respect to the accuracy of random forest, averaged over all datasets. (b) Corresponding p-values of a set of post-hoc Nemenyi tests. Green shows significant differences at 0.01, yellow significant differences at 0.05.}
    \label{fig:stat_rf}
\end{figure}

In Fig. \ref{fig:stat_rmse} we provide the average rankings and p-values for the statistical test performed on the results of Section 5.1 of the paper (Nemenyi post-hoc tests on all pairs of algorithms), corresponding to Tab. \ref{tab:rmse}. In Fig. \ref{fig:stat_rf} we instead provide average rankings and corresponding p-values for the statistical tests performed on the random forest classifier of Section 5.2, corresponding to Tab. \ref{tab:rf}. The results are discussed more in-depth in the main paper in Section 5.2.

\subsection*{Ablation study with a computational budget (Section 5.3)}

\begin{figure}
    \centering
    \subfigure[]{\includegraphics[width=0.59\columnwidth]{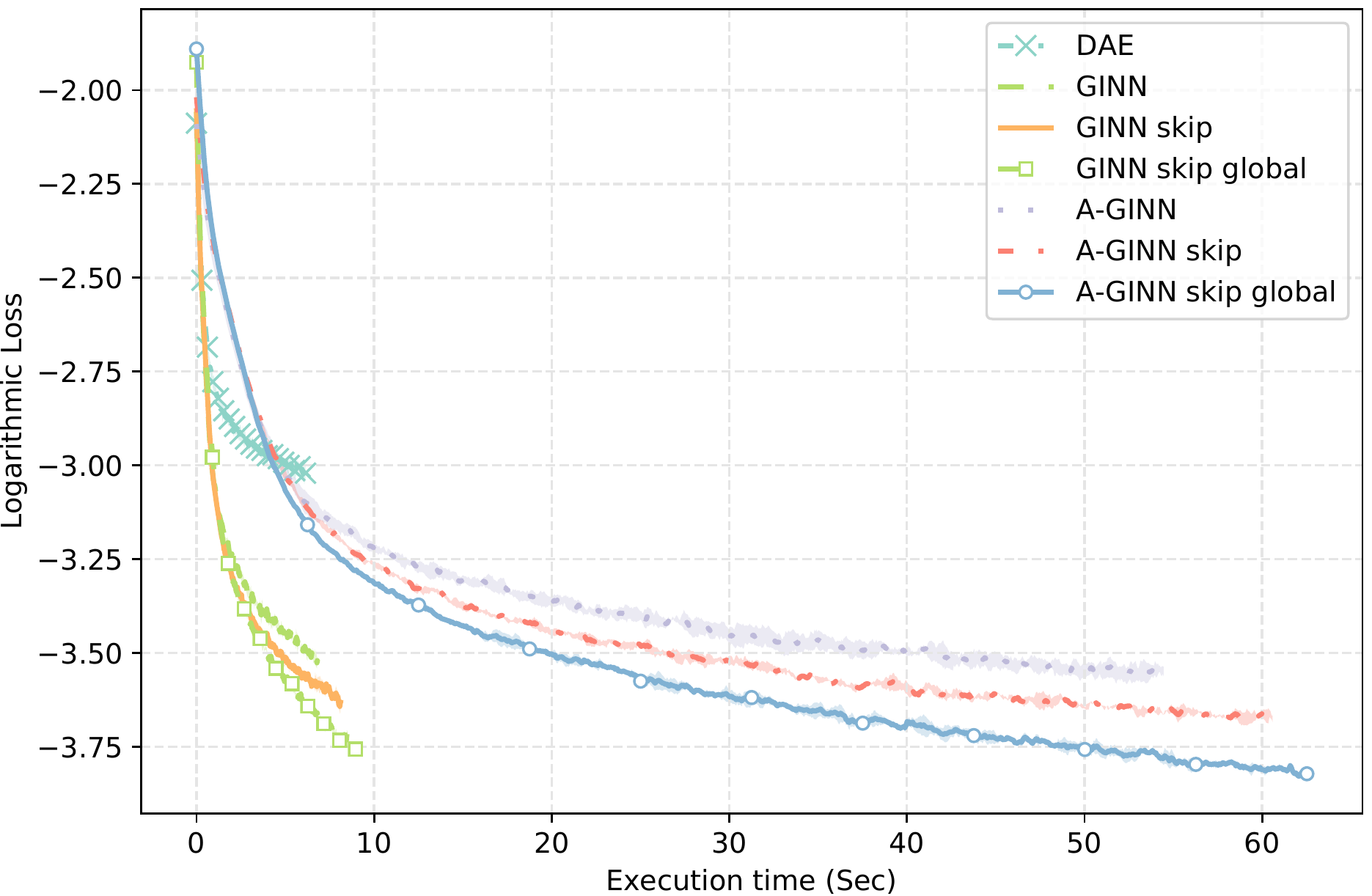}}
    \subfigure[]{\includegraphics[width=0.59\columnwidth]{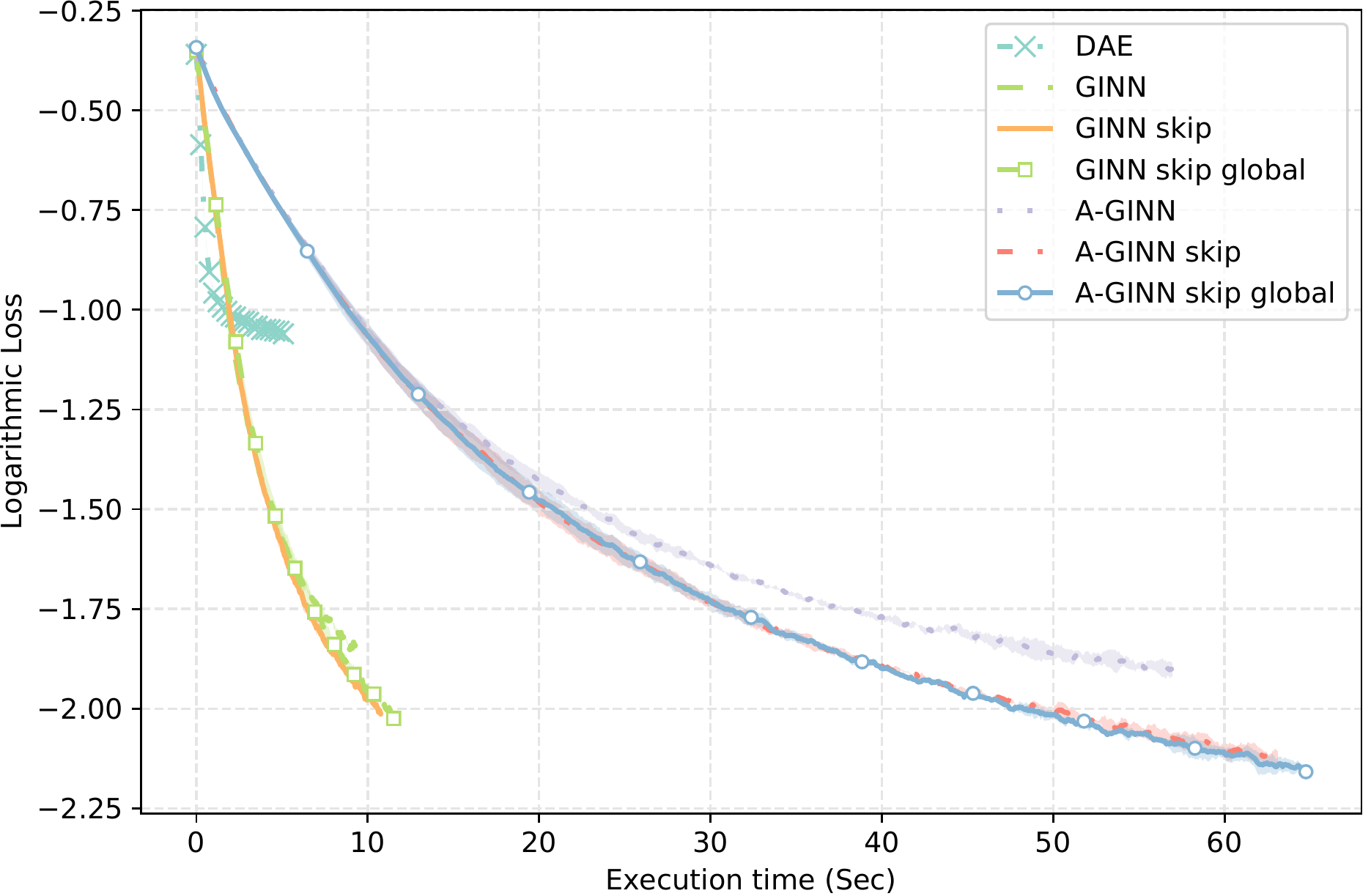}}
    \subfigure[]{\includegraphics[width=0.59\columnwidth]{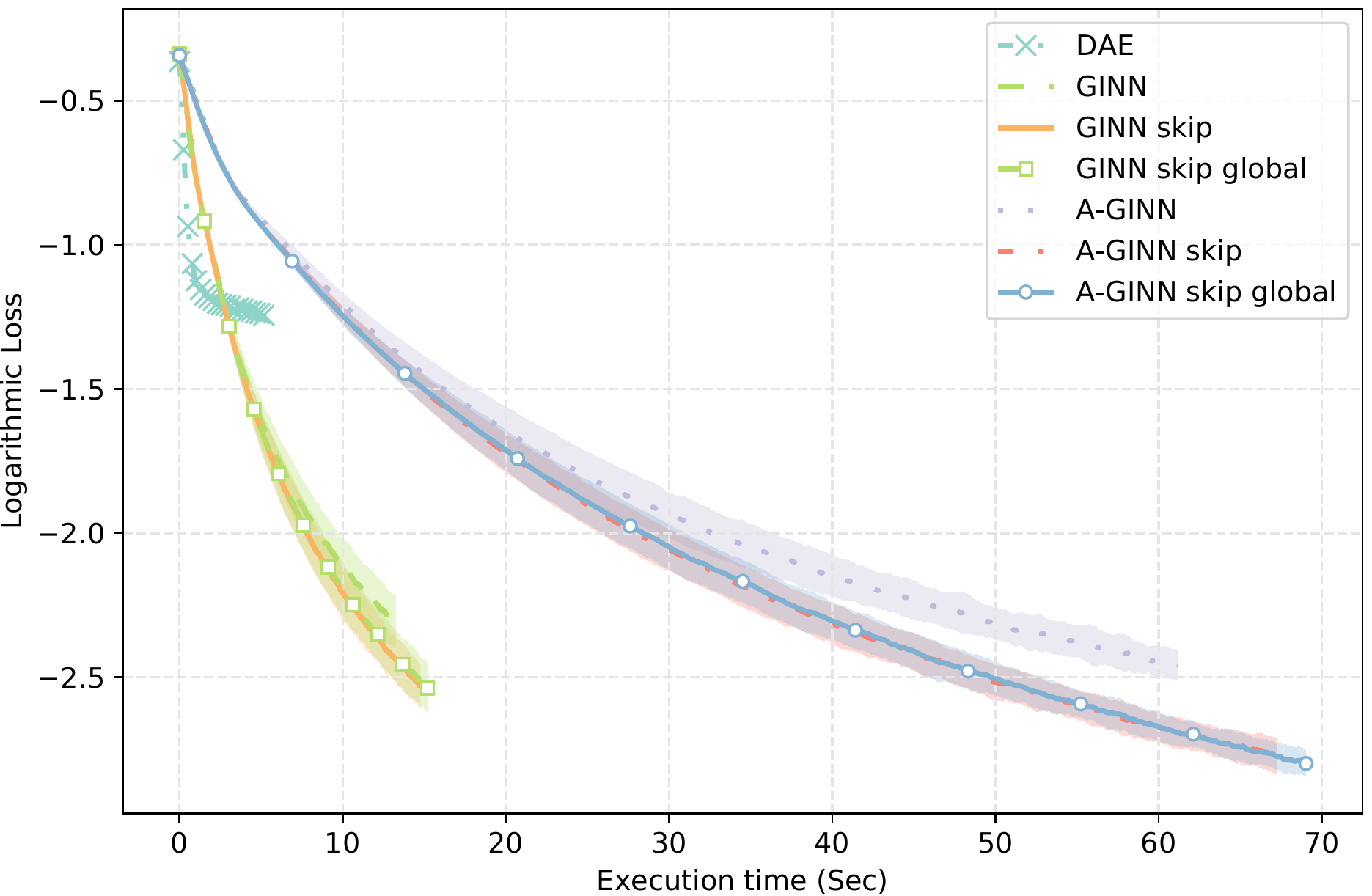}}
    \caption{Convergence of the logarithmic loss function with respect to a fixed computational budget on the Ionosphere (a), Tic-Tac-Toe (b) and Phishing (c) datasets for each variant described in the ablation study of Section 5.3.}
    \label{fig:ablation_convergence_with_budget}
\end{figure}

In this section we replicate the ablation study of Section 5.3, but we evaluate the convergence of each variant with respect to a fixed computational budget, as shown in Fig. \ref{fig:ablation_convergence_with_budget}. As can be seen, our baseline method with GCN but no adversarial loss can converge to a significantly better result than a standard DAE in a fraction of the time. Including the adversarial loss slightly improves the final result, requiring however $3-4$ times the budget of the baseline method.

\subsection*{Detailed results for the evaluation on real-world datasets}

\begin{figure}[p]
	\centering
	\subfigure[Mammographic mass]{\includegraphics[width=0.8\textwidth]{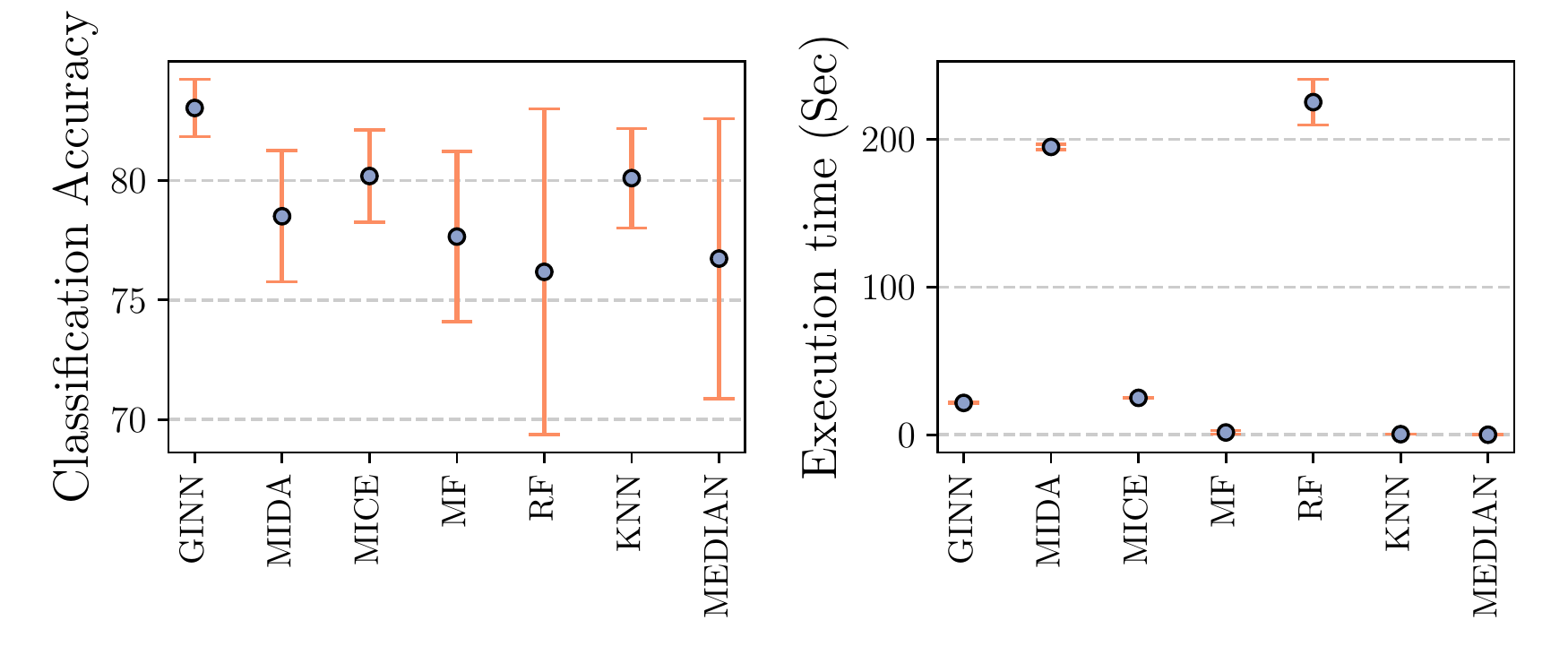}}
	\subfigure[Cervical cancer]{\includegraphics[width=0.8\textwidth]{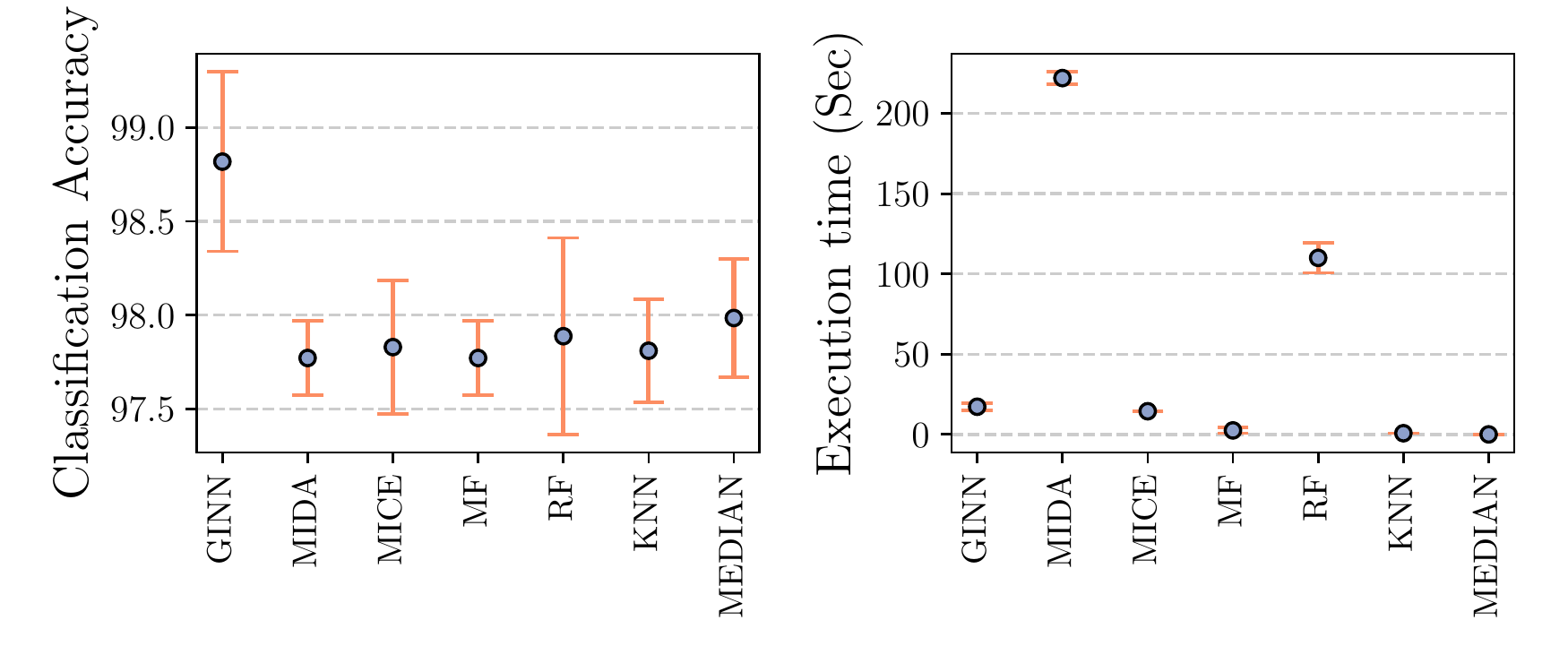}}
	\subfigure[Air quality]{\includegraphics[width=0.8\textwidth]{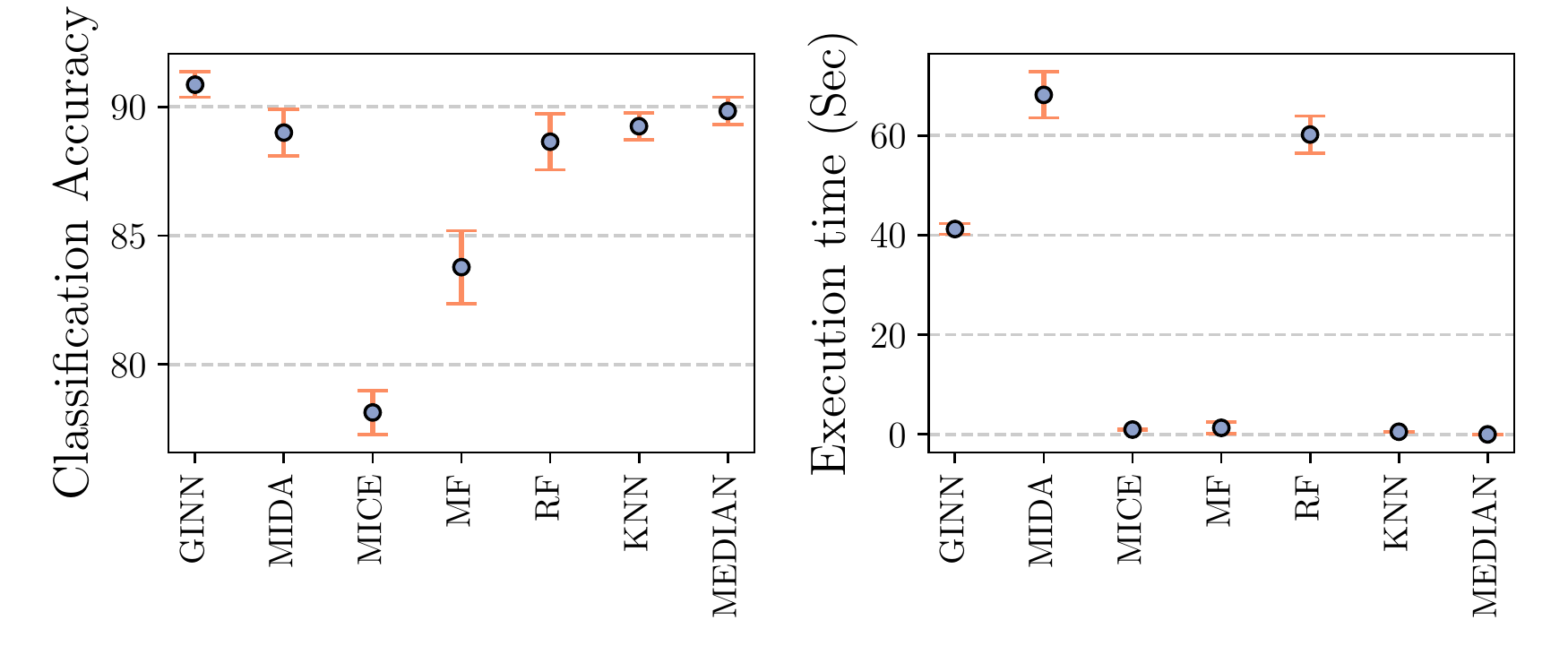}}
	\caption{Random forest classification accuracy and imputation times on (a) Mammographic mass  (b) Cervical cancer and (c) Air quality datasets.}
	\label{fig:real-missing_new}
\end{figure}

\begin{figure}
    \centering
    \subfigure[Cervical cancer]{\includegraphics[width=0.45\columnwidth]{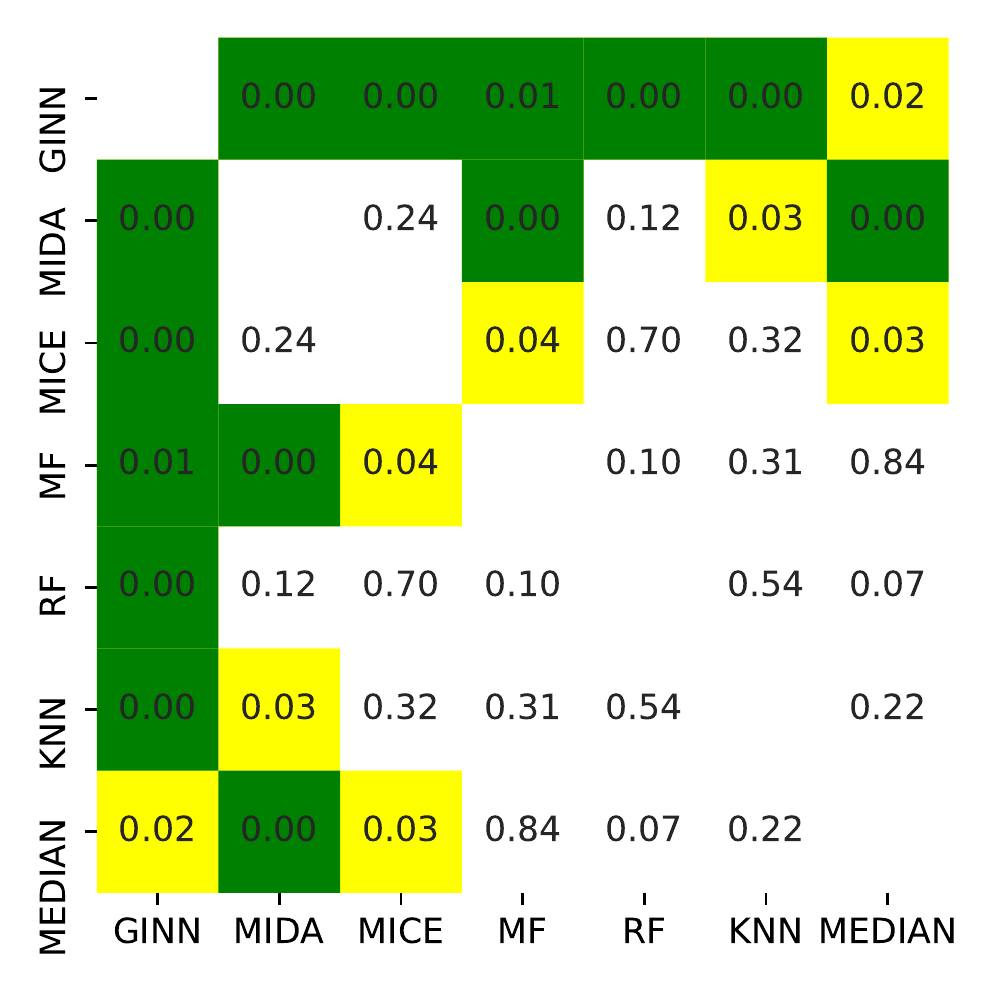}}
    \subfigure[Mammography mass]{\includegraphics[width=0.45\columnwidth]{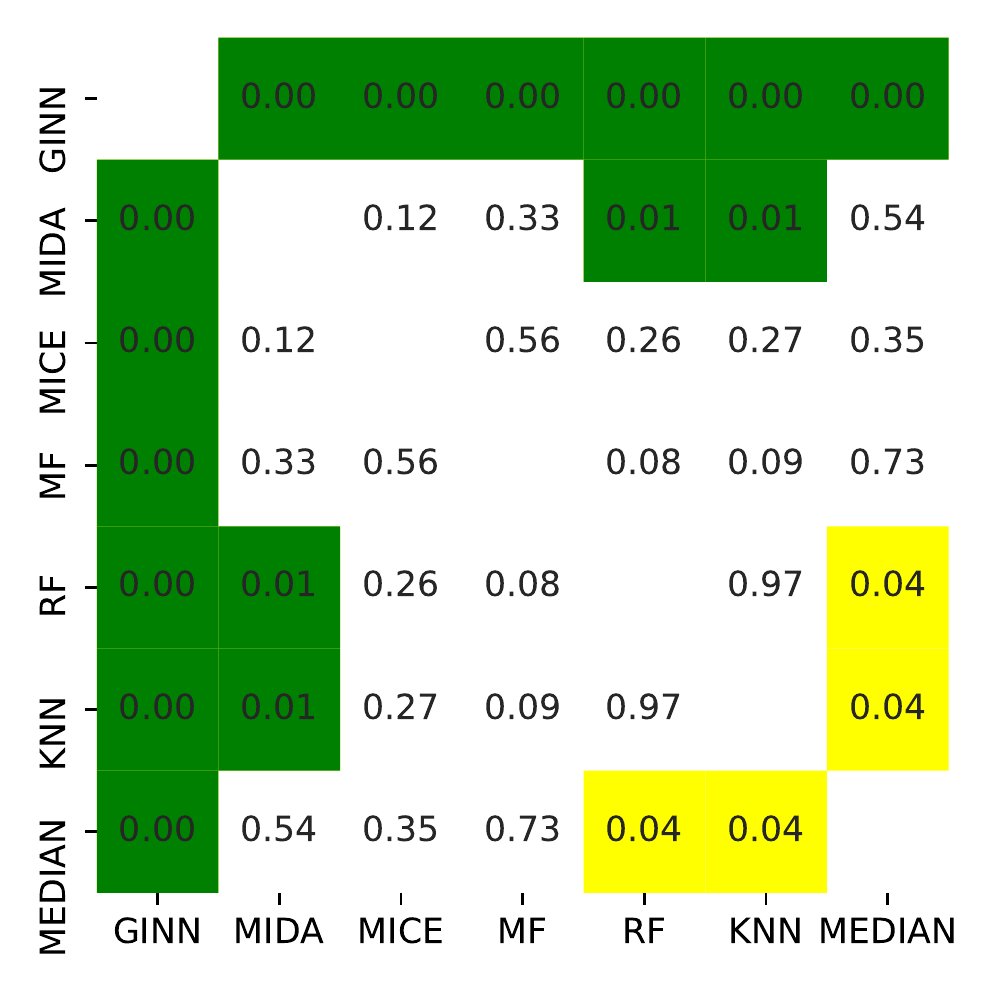}}
    \subfigure[Air quality]{\includegraphics[width=0.45\columnwidth]{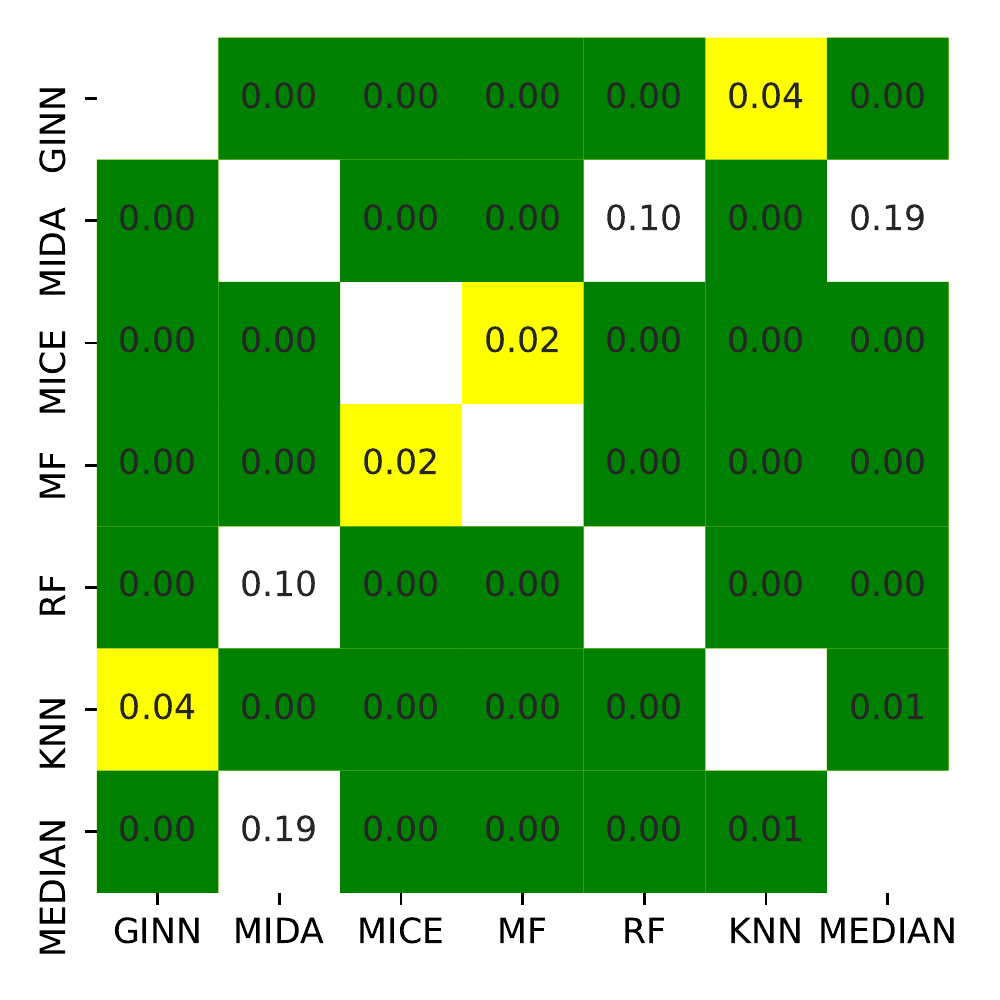}}
    \caption{Corresponding p-values of a set of post-hoc Nemenyi tests on the rankings of Table 6 in the paper relative to the downstream classification/regression tasks on the three datasets with pre-existing missing data: (a) Cervical cancer (13\% of missing elements), (b) Mammographic mass  (4\% of missing elements) and (c) Air quality  (13\% of missing elements). Green shows significant differences at 0.01, yellow significant differences at 0.05.}
    \label{fig:stat}
\end{figure}

{In Figure \ref{fig:real-missing_new} we report the scores obtained with a random forest classifier and the time in seconds needed for the imputation, including for GINN, the time needed for the similarity-graph construction. As can be seen, GINN's imputation allows to achieve the best accuracy and consistency across all three datasets, even in the case where the imputed training labels are used to train the classification model. Regarding execution times, GINN is considerably faster than alternative neural approaches and missForest, but it is slower than the approaches that do not perform a training phase.

We corroborate the results by performing a statistical analysis. To focus on the impact of the imputation techniques, we compute the relative ranking for each combination of imputation method and classification algorithm, and we then aggregate the results with respect to the latter.} A Friedman rank test confirms that there are statistical significant differences with respect to the accuracy of the classifiers with p-values of $1.6e^{-7}$, $1.3e^{-7}$, $1.2e^{-9}$ respectively for the Cervical cancer, Mammographic mass and Air quality datasets. A successive set of Nemenyi post-hoc tests, between all pairs of algorithms, further confirms statistical significant differences between GINN and all other methods as reported in Figure \ref{fig:stat}.

\pagebreak

\footnotesize{
\begin{longtable}{lccccccc}
\hline
RMSE & GINN & MIDA & MICE & MF & RF & KNN & MEDIAN \\
\hline
\endhead
\hline
\endfoot
abalone\_10 & 0.901 & 0.987 & 1.002 & 3.110 & 0.693 & 0.741 & 1.031 \\
abalone\_20 & 0.919 & 1.087 & 1.031 & 3.335 & 0.783 & 1.031 & 1.164 \\
abalone\_30 & 0.904 & 1.190 & 1.069 & 3.321 & 0.811 & 1.072 & 1.112 \\
abalone\_50 & 1.038 & 1.218 & 1.052 & 2.679 & 0.948 & 1.121 & 1.155 \\
anuran-calls\_10 & 0.066 & 0.156 & 0.069 & 0.745 & 0.159 & 0.049 & 0.225 \\
anuran-calls\_20 & 0.067 & 0.157 & 0.073 & 0.750 & 0.158 & 0.052 & 0.221 \\
anuran-calls\_30 & 0.070 & 0.187 & 0.076 & 0.091 & 0.154 & 0.058 & 0.226 \\
anuran-calls\_50 & 0.081 & 0.202 & 0.086 & 0.097 & 0.132 & 0.089 & 0.227 \\
balance-scale\_10 & 0.565 & 0.424 & 0.579 & 0.545 & 0.552 & 0.614 & 0.591 \\
balance-scale\_20 & 0.575 & 0.426 & 0.559 & 0.540 & 0.558 & 0.604 & 0.586 \\
balance-scale\_30 & 0.559 & 0.436 & 0.579 & 0.521 & 0.516 & 0.580 & 0.577 \\
balance-scale\_50 & 0.548 & 0.441 & 0.566 & 0.495 & 0.482 & 0.581 & 0.577 \\
breast-cancer\_10 & 45.901 & 65.698 & 35.837 & 25.006 & 6.021 & 28.299 & 105.975 \\
breast-cancer\_20 & 39.258 & 84.307 & 35.248 & 23.413 & 19.332 & 27.415 & 122.510 \\
breast-cancer\_30 & 54.633 & 114.390 & 38.122 & 23.559 & 21.155 & 39.653 & 127.645 \\
breast-cancer\_50 & 57.873 & 136.427 & 37.325 & 26.060 & 28.434 & 48.219 & 121.975 \\
car-evaluation \_10 & 0.616 & 0.621 & 0.645 & 0.846 & 0.631 & 0.660 & 0.653 \\
car-evaluation \_20 & 0.613 & 0.619 & 0.628 & 0.845 & 0.637 & 0.647 & 0.646 \\
car-evaluation \_30 & 0.623 & 0.632 & 0.636 & 0.845 & 0.636 & 0.636 & 0.649 \\
car-evaluation \_50 & 0.615 & 0.629 & 0.641 & 0.846 & 0.636 & 0.648 & 0.640 \\
credit-card\_10 & 17585.4 & 18323.0 & 16034.1 & 15863.5 & 11762.4 & 15939.1 & 24680.3 \\
credit-card\_20 & 17792.4 & 18462.5 & 15848.6 & 14864.0 & 12055.4 & 15907.4 & 23884.7 \\
credit-card\_30 & 17479.4 & 19664.1 & 15687.0 & 15212.9 & 12897.4 & 17294.4 & 23708.0 \\
credit-card\_50 & 19437.2 & 20063.7 & 16394.4 & 15471.8 & 14139.7 & 20974.4 & 23652.4 \\
electrical-grid\_10 & 1.493 & 1.871 & 1.625 & 1.375 & 1.398 & 1.586 & 1.574 \\
electrical-grid\_20 & 1.494 & 1.855 & 1.638 & 1.318 & 1.438 & 1.624 & 1.538 \\
electrical-grid\_30 & 1.534 & 1.857 & 1.637 & 1.474 & 1.536 & 1.763 & 1.555 \\
electrical-grid\_50 & 1.570 & 1.878 & 1.656 & 1.686 & 1.724 & 1.666 & 1.580 \\
heart\_10 & 9.125 & 17.100 & 9.696 & 9.901 & 9.869 & 10.477 & 9.553 \\
heart\_20 & 8.068 & 18.842 & 11.220 & 11.363 & 10.046 & 12.816 & 10.900 \\
heart\_30 & 10.883 & 25.212 & 12.326 & 11.367 & 11.368 & 14.846 & 11.708 \\
heart\_50 & 10.301 & 20.874 & 11.671 & 10.743 & 10.910 & 11.651 & 11.004 \\
ionosphere\_10 & 0.364 & 0.772 & 0.416 & 0.572 & 0.349 & 0.348 & 0.527 \\
ionosphere\_20 & 0.385 & 0.819 & 0.458 & 351.366 & 0.386 & 0.379 & 0.542 \\
ionosphere\_30 & 0.425 & 1.118 & 0.462 & 340.257 & 0.407 & 0.380 & 0.551 \\
ionosphere\_50 & 0.442 & 1.170 & 0.475 & 185.546 & 0.431 & 0.415 & 0.544 \\
iris\_10 & 0.400 & 1.988 & 0.386 & 0.116 & 0.237 & 0.287 & 0.895 \\
iris\_20 & 0.384 & 1.562 & 0.411 & 0.133 & 0.310 & 0.352 & 1.261 \\
iris\_30 & 0.349 & 1.792 & 0.379 & 0.106 & 0.333 & 0.446 & 1.159 \\
iris\_50 & 0.385 & 1.992 & 0.453 & 0.109 & 0.534 & 0.414 & 1.135 \\
page-blocks\_10 & 407.7 & 814.3 & 1511.9 & 684.5 & 206.7 & 241.6 & 1019.8 \\
page-blocks\_20 & 780.0 & 2305.8 & 2006.4 & 1349.1 & 934.1 & 723.1 & 2512.4 \\
page-blocks\_30 & 536.6 & 1035.7 & 1889.0 & 795.4 & 579.2 & 604.1 & 869.4 \\
page-blocks\_50 & 1438.4 & 1873.5 & 1624.9 & 1044.6 & 1291.9 & 1858.2 & 1801.9 \\
phishing\_10 & 0.500 & 0.506 & 0.618 & 0.520 & 0.607 & 0.519 & 0.576 \\
phishing\_20 & 0.513 & 0.514 & 0.623 & 0.531 & 0.601 & 0.524 & 0.579 \\
phishing\_30 & 0.493 & 0.529 & 0.625 & 0.594 & 0.606 & 0.548 & 0.581 \\
phishing\_50 & 0.506 & 0.541 & 0.623 & 0.581 & 0.596 & 0.538 & 0.576 \\
satellite\_10 & 5.978 & 9.251 & 4.162 & 4.846 & 3.207 & 4.077 & 18.293 \\
satellite\_20 & 6.066 & 9.802 & 4.312 & 4.158 & 3.404 & 4.257 & 18.423 \\
satellite\_30 & 6.256 & 11.528 & 4.481 & 4.562 & 3.634 & 4.523 & 18.335 \\
satellite\_50 & 6.934 & 14.146 & 5.139 & 5.179 & 4.226 & 6.255 & 18.375 \\
tic-tac-toe\_10 & 0.467 & 0.512 & 0.665 & 0.816 & 0.656 & 0.625 & 0.624 \\
tic-tac-toe\_20 & 0.466 & 0.527 & 0.662 & 0.816 & 0.657 & 0.616 & 0.614 \\
tic-tac-toe\_30 & 0.469 & 0.576 & 0.657 & 0.816 & 0.671 & 0.625 & 0.622 \\
tic-tac-toe\_50 & 0.470 & 0.606 & 0.661 & 0.816 & 0.685 & 0.647 & 0.618 \\
student\_10 & 0.258 & 0.247 & 0.897 & 0.897 & 0.293 & 0.293 & 0.514 \\
student\_20 & 0.258 & 0.247 & 0.536 & 0.288 & 0.289 & 0.294 & 0.514 \\
student\_30 & 0.295 & 0.285 & 0.537 & 0.287 & 0.292 & 0.303 & 0.515 \\
student\_50 & 0.298 & 0.268 & 0.536 & 0.307 & 0.289 & 0.322 & 0.515 \\
wine\_10 & 43.473 & 190.147 & 56.868 & 51.250 & 54.981 & 44.689 & 121.008 \\
wine\_20 & 46.014 & 171.845 & 46.775 & 46.133 & 51.354 & 53.893 & 106.947 \\
wine\_30 & 47.901 & 154.344 & 52.358 & 54.764 & 48.672 & 54.636 & 99.651 \\
wine\_50 & 49.698 & 133.361 & 55.972 & 51.696 & 54.857 & 63.823 & 82.517 \\
wine-red\_10 & 6.305 & 8.690 & 7.017 & 7.991 & 5.181 & 5.951 & 8.889 \\
wine-red\_20 & 7.244 & 9.458 & 8.143 & 7.287 & 7.531 & 7.316 & 9.586 \\
wine-red\_30 & 8.099 & 10.313 & 9.092 & 8.883 & 8.414 & 9.752 & 10.265 \\
wine-red\_50 & 8.381 & 10.114 & 9.566 & 9.636 & 10.265 & 10.408 & 9.879 \\
wine-white\_10 & 10.115 & 13.103 & 10.551 & 39.086 & 8.009 & 8.817 & 12.908 \\
wine-white\_20 & 10.783 & 14.253 & 11.387 & 42.724 & 9.126 & 10.923 & 13.902 \\
wine-white\_30 & 11.188 & 15.383 & 11.174 & 41.324 & 10.350 & 12.454 & 13.432 \\
wine-white\_50 & 11.214 & 15.699 & 12.028 & 39.693 & 11.935 & 14.184 & 13.373 \\
wireless\_10 & 4.438 & 6.707 & 4.452 & 5.165 & 3.677 & 3.863 & 7.967 \\
wireless\_20 & 4.735 & 7.082 & 4.554 & 2.325 & 3.892 & 4.491 & 8.464 \\
wireless\_30 & 4.851 & 8.470 & 4.728 & 1.863 & 4.214 & 4.995 & 8.588 \\
wireless\_50 & 4.473 & 9.132 & 4.677 & 4.918 & 4.496 & 4.629 & 8.470 \\
yeast\_10 & 0.084 & 0.131 & 0.091 & 0.109 & 0.088 & 0.088 & 0.110 \\
yeast\_20 & 0.083 & 0.126 & 0.089 & 0.082 & 0.087 & 0.091 & 0.099 \\
yeast\_30 & 0.085 & 0.162 & 0.090 & 0.093 & 0.086 & 0.091 & 0.102 \\
yeast\_50 & 0.088 & 0.156 & 0.096 & 0.088 & 0.096 & 0.092 & 0.102 \\
\hline
\caption{Root Mean Squared Error (RMSE) of the imputation performance for all considered algorithms with respect to all possible percentages of missing elements.}
\label{tab:rmse_additional}
\end{longtable}
}
\pagebreak

\footnotesize{
\begin{longtable}{lcccccccc}
\hline
Random forest & GT & GINN & MIDA & MICE & MF & RF & KNN & MEDIAN\\
\hline
\endhead
\hline
\endfoot
abalone\_10 & 52.632 & 55.263 & 52.871 & 53.190 & 55.024 & 55.343 & 54.147 & 53.907 \\
abalone\_20 & 53.748 & 54.785 & 53.030 & 53.907 & 52.791 & 52.951 & 55.263 & 52.313 \\
abalone\_30 & 53.270 & 55.821 & 53.270 & 54.944 & 51.754 & 52.472 & 53.030 & 51.675 \\
abalone\_50 & 54.545 & 54.625 & 50.638 & 52.233 & 52.313 & 53.987 & 53.270 & 53.828 \\
anuran-calls\_10 & 93.191 & 93.654 & 92.450 & 91.292 & 90.968 & 92.867 & 92.265 & 91.292 \\
anuran-calls\_20 & 93.840 & 93.747 & 90.690 & 92.774 & 90.968 & 92.497 & 93.747 & 91.014 \\
anuran-calls\_30 & 92.728 & 94.442 & 89.810 & 91.848 & 93.191 & 92.635 & 91.061 & 90.412 \\
anuran-calls\_50 & 93.654 & 92.913 & 89.208 & 91.431 & 91.570 & 93.793 & 92.635 & 90.366 \\
balance-scale\_10 & 78.191 & 79.787 & 75.532 & 74.468 & 80.319 & 79.255 & 71.809 & 77.128 \\
balance-scale\_20 & 79.255 & 79.787 & 78.191 & 77.128 & 71.277 & 73.936 & 75.532 & 76.596 \\
balance-scale\_30 & 81.383 & 76.596 & 72.340 & 76.596 & 77.660 & 76.064 & 78.723 & 69.149 \\
balance-scale\_50 & 80.851 & 75.532 & 63.830 & 61.702 & 72.872 & 71.809 & 75.000 & 65.957 \\
breast-cancer\_10 & 97.661 & 95.322 & 97.661 & 95.906 & 95.906 & 98.246 & 95.906 & 95.322 \\
breast-cancer\_20 & 96.491 & 96.491 & 95.322 & 96.491 & 95.906 & 94.152 & 96.491 & 95.906 \\
breast-cancer\_30 & 97.076 & 96.491 & 95.906 & 94.737 & 94.152 & 97.076 & 96.491 & 97.661 \\
breast-cancer\_50 & 96.491 & 94.737 & 90.058 & 94.737 & 94.737 & 96.491 & 96.491 & 93.567 \\
car-evaluation \_10 & 71.291 & 71.484 & 71.098 & 70.135 & 69.942 & 69.942 & 70.520 & 70.135 \\
car-evaluation \_20 & 70.328 & 71.869 & 70.328 & 69.942 & 69.942 & 71.676 & 69.942 & 70.135 \\
car-evaluation \_30 & 70.520 & 72.832 & 71.484 & 69.942 & 69.942 & 70.713 & 70.520 & 69.942 \\
car-evaluation \_50 & 73.218 & 71.869 & 73.603 & 69.942 & 70.906 & 71.484 & 71.484 & 71.869 \\
credit-card\_10 & 77.933 & 78.453 & 77.887 & 77.873 & 78.160 & 77.947 & 77.880 & 77.893 \\
credit-card\_20 & 77.913 & 77.913 & 77.907 & 77.880 & 77.907 & 77.873 & 77.853 & 77.973 \\
credit-card\_30 & 77.873 & 77.980 & 77.880 & 77.880 & 77.887 & 77.893 & 77.900 & 77.900 \\
credit-card\_50 & 78.433 & 78.933 & 78.367 & 77.880 & 77.933 & 77.887 & 77.893 & 77.920 \\
electrical-grid\_10 & 97.200 & 97.433 & 95.600 & 99.667 & 97.433 & 88.733 & 99.933 & 98.867 \\
electrical-grid\_20 & 99.933 & 99.567 & 98.633 & 93.267 & 99.533 & 94.500 & 94.667 & 99.500 \\
electrical-grid\_30 & 98.533 & 91.100 & 86.633 & 97.567 & 94.900 & 97.867 & 98.700 & 99.367 \\
electrical-grid\_50 & 95.633 & 93.900 & 83.933 & 95.633 & 91.433 & 93.467 & 99.867 & 99.367 \\
heart\_10 & 81.319 & 84.615 & 82.418 & 80.220 & 83.516 & 78.022 & 73.626 & 82.418 \\
heart\_20 & 80.220 & 85.714 & 80.220 & 79.121 & 78.022 & 85.714 & 81.319 & 78.022 \\
heart\_30 & 82.418 & 83.516 & 73.626 & 83.516 & 78.022 & 82.418 & 82.418 & 76.923 \\
heart\_50 & 78.022 & 76.923 & 74.725 & 76.923 & 76.923 & 82.418 & 80.220 & 73.626 \\
ionosphere\_10 & 90.566 & 90.566 & 92.453 & 92.453 & 90.566 & 91.509 & 91.509 & 91.509 \\
ionosphere\_20 & 94.340 & 91.509 & 86.792 & 92.453 & 88.679 & 92.453 & 91.509 & 93.396 \\
ionosphere\_30 & 90.566 & 95.283 & 84.906 & 92.453 & 92.453 & 92.453 & 95.283 & 93.396 \\
ionosphere\_50 & 92.453 & 92.453 & 78.302 & 91.509 & 86.792 & 91.509 & 91.509 & 92.453 \\
iris\_10 & 88.889 & 95.556 & 93.333 & 93.333 & 91.111 & 93.333 & 93.333 & 95.556 \\
iris\_20 & 91.111 & 91.111 & 91.111 & 88.889 & 91.111 & 91.111 & 93.333 & 91.111 \\
iris\_30 & 91.111 & 88.889 & 80.000 & 93.333 & 91.111 & 91.111 & 93.333 & 86.667 \\
iris\_50 & 93.333 & 88.889 & 88.889 & 88.889 & 91.111 & 88.889 & 91.111 & 84.444 \\
page-blocks\_10 & 94.762 & 95.250 & 94.458 & 94.093 & 94.580 & 94.884 & 94.580 & 95.128 \\
page-blocks\_20 & 94.153 & 94.884 & 95.250 & 93.910 & 94.093 & 95.371 & 95.371 & 94.945 \\
page-blocks\_30 & 94.641 & 95.067 & 94.823 & 94.762 & 94.641 & 94.641 & 94.884 & 94.702 \\
page-blocks\_50 & 94.458 & 95.615 & 94.032 & 92.387 & 94.641 & 94.702 & 95.067 & 94.823 \\
phishing\_10 & 84.236 & 86.453 & 86.207 & 83.744 & 84.483 & 84.729 & 85.222 & 83.251 \\
phishing\_20 & 85.222 & 84.975 & 85.961 & 85.714 & 83.005 & 84.483 & 83.744 & 84.975 \\
phishing\_30 & 84.729 & 84.483 & 83.005 & 83.251 & 83.990 & 83.005 & 84.236 & 82.020 \\
phishing\_50 & 83.251 & 85.468 & 80.296 & 80.542 & 82.759 & 82.512 & 79.310 & 80.542 \\
satellite\_10 & 83.450 & 82.550 & 82.850 & 83.350 & 82.750 & 82.000 & 82.550 & 80.650 \\
satellite\_20 & 83.400 & 83.300 & 82.400 & 82.150 & 82.050 & 82.650 & 83.250 & 79.700 \\
satellite\_30 & 83.900 & 83.850 & 81.700 & 82.550 & 83.650 & 83.300 & 83.500 & 79.950 \\
satellite\_50 & 83.200 & 82.850 & 78.150 & 83.250 & 81.800 & 83.050 & 83.350 & 79.650 \\
tic-tac-toe\_10 & 74.306 & 75.000 & 70.833 & 67.708 & 68.056 & 67.708 & 69.792 & 68.056 \\
tic-tac-toe\_20 & 69.097 & 70.486 & 70.486 & 66.667 & 71.875 & 65.972 & 70.833 & 72.222 \\
tic-tac-toe\_30 & 69.444 & 72.569 & 68.056 & 67.708 & 67.361 & 65.972 & 71.181 & 72.917 \\
tic-tac-toe\_50 & 68.403 & 68.750 & 69.444 & 65.625 & 65.625 & 68.750 & 65.972 & 66.319 \\
student\_10 & 83.104 & 85.338 & 82.188 & 83.505 & 80.928 & 82.245 & 85.052 & 83.448 \\
student\_20 & 82.245 & 82.188 & 84.364 & 83.505 & 84.364 & 84.021 & 81.787 & 83.104 \\
student\_30 & 84.021 & 84.651 & 84.593 & 81.214 & 84.651 & 83.276 & 83.505 & 83.219 \\
student\_50 & 82.761 & 83.333 & 82.245 & 85.739 & 82.188 & 84.135 & 84.593 & 84.021 \\
wine\_10 & 92.593 & 94.444 & 94.444 & 98.148 & 96.296 & 98.148 & 94.444 & 100.000 \\
wine\_20 & 92.593 & 94.444 & 94.444 & 96.296 & 98.148 & 90.741 & 92.593 & 98.148 \\
wine\_30 & 96.296 & 98.148 & 92.593 & 98.148 & 94.444 & 98.148 & 94.444 & 94.444 \\
wine\_50 & 90.741 & 98.148 & 75.926 & 98.148 & 92.593 & 92.593 & 92.593 & 94.444 \\
wine-red\_10 & 61.875 & 63.958 & 59.375 & 60.417 & 61.458 & 63.750 & 62.917 & 61.875 \\
wine-red\_20 & 61.875 & 62.917 & 60.417 & 60.833 & 59.375 & 58.542 & 62.917 & 59.792 \\
wine-red\_30 & 63.333 & 62.708 & 62.083 & 61.667 & 63.333 & 62.292 & 61.458 & 58.125 \\
wine-red\_50 & 63.542 & 58.542 & 53.958 & 60.625 & 58.542 & 59.792 & 55.000 & 52.917 \\
wine-white\_10 & 51.565 & 54.490 & 52.585 & 53.401 & 51.565 & 53.469 & 53.469 & 53.061 \\
wine-white\_20 & 52.789 & 54.490 & 51.973 & 54.626 & 50.816 & 52.517 & 54.082 & 53.946 \\
wine-white\_30 & 52.109 & 53.673 & 51.361 & 52.517 & 52.313 & 53.265 & 51.020 & 51.837 \\
wine-white\_50 & 54.422 & 53.401 & 50.272 & 52.517 & 49.728 & 51.156 & 50.272 & 50.340 \\
wireless\_10 & 97.500 & 97.667 & 97.333 & 97.667 & 97.667 & 97.000 & 97.000 & 97.667 \\
wireless\_20 & 96.833 & 97.000 & 97.000 & 93.667 & 96.000 & 94.333 & 95.500 & 95.833 \\
wireless\_30 & 97.167 & 96.333 & 93.167 & 97.333 & 96.667 & 97.000 & 95.833 & 97.167 \\
wireless\_50 & 96.500 & 97.500 & 95.333 & 93.333 & 95.333 & 96.000 & 97.167 & 95.833 \\
yeast\_10 & 57.848 & 59.193 & 59.193 & 58.072 & 54.709 & 59.193 & 56.951 & 57.399 \\
yeast\_20 & 58.744 & 61.211 & 55.381 & 59.865 & 57.623 & 60.090 & 58.744 & 56.726 \\
yeast\_30 & 55.157 & 56.502 & 46.188 & 58.072 & 43.722 & 54.484 & 56.502 & 53.363 \\
yeast\_50 & 57.623 & 49.327 & 39.910 & 56.951 & 47.309 & 52.691 & 54.036 & 56.278 \\
\hline
\caption{Classification accuracy on each dataset using a random forest classifier. The model was trained over the imputed data by the analyzed algorithms for all percentage of missing elements.}
\label{tab:rf}
\end{longtable}
}
\pagebreak

\end{document}